\documentclass[sigconf]{acmart}
\usepackage{hyperref}
\usepackage{graphicx}
\usepackage{caption}
\usepackage{subcaption}

\usepackage{xcolor}
\usepackage{enumitem}
\usepackage{bbm}
\usepackage{multirow}
\usepackage{amsmath,amsfonts}
\usepackage{textcomp}
\usepackage{diagbox}
\usepackage{xspace}
\usepackage{algorithm}
\usepackage{algpseudocode}
\usepackage{url}
\usepackage{balance}
\newcommand{\bs}{\boldsymbol}
\newcommand{\s}{\bs{s}}
\newcommand{\btheta}{\bs{\theta}}

\newcommand{\IACC}{IaC\textsuperscript{C}\xspace}
\newcommand{\IACT}{IaC\textsuperscript{T}\xspace}
\newcommand{\IAC}{IaC\xspace}

\newcommand{\minisection}[1]{\vspace{5pt}\noindent\textbf{#1.}}
\DeclareMathOperator*{\argmax}{arg\,max}

\AtBeginDocument{%
  \providecommand\BibTeX{{%
    \normalfont B\kern-0.5em{\scshape i\kern-0.25em b}\kern-0.8em\TeX}}}

\copyrightyear{2023} 
\acmYear{2023} 
\setcopyright{acmlicensed}\acmConference[KDD '23]{Proceedings of the 29th ACM SIGKDD Conference on Knowledge Discovery and Data Mining}{August 6--10, 2023}{Long Beach, CA, USA}
\acmBooktitle{Proceedings of the 29th ACM SIGKDD Conference on Knowledge Discovery and Data Mining (KDD '23), August 6--10, 2023, Long Beach, CA, USA}
\acmPrice{15.00}
\acmDOI{10.1145/3580305.3599856}
\acmISBN{979-8-4007-0103-0/23/08}

\begin{document}

\title{Learning Multi-Agent Intention-Aware Communication for Optimal Multi-Order Execution in Finance}

\author{Yuchen Fang}
\authornote{These authors contributed equally to this research.}
\authornote{This work was conducted during the internship of Yuchen Fang and Zhenggang Tang at Microsoft Research Asia.}
\email{arthur\_fyc@sjtu.edu.cn}
\affiliation{
    \institution{Shanghai Jiao Tong University}
    \country{}
}

\author{Zhenggang Tang}
\authornotemark[1]
\authornotemark[2]
\email{zt15@illinois.edu}
\affiliation{
    \institution{University of Illinois Urbana-Champaign}
    \country{}
}

\author{Kan Ren}
\authornotemark[1]
\authornote{Corresponding author.}
\email{kan.ren@microsoft.com}
\affiliation{
    \institution{Microsoft Research Asia}
    \country{}
}

\author{Weiqing Liu}
\affiliation{
    \institution{Microsoft Research Asia}
    \country{}
}

\author{Li Zhao}
\affiliation{
    \institution{Microsoft Research Asia}
    \country{}
}
\author{Jiang Bian}
\affiliation{
    \institution{Microsoft Research Asia}
    \country{}
}

\author{Dongsheng Li}
\affiliation{
    \institution{Microsoft Research Asia}
    \country{}
}

\author{Weinan Zhang}
\authornotemark[3]
\email{wnzhang@sjtu.edu.cn}
\affiliation{
    \institution{Shanghai Jiao Tong University}
    \country{}
}

\author{Yong Yu}
\affiliation{
    \institution{Shanghai Jiao Tong University}
    \country{}
}

\author{Tie-Yan Liu}
\affiliation{
    \institution{Microsoft Research Asia}
    \country{}
}

\renewcommand{\shortauthors}{Yuchen Fang et al.}

\begin{abstract}
Order execution is a fundamental task in quantitative finance, aiming at finishing acquisition or liquidation for a number of trading orders of the specific assets. Recent advance in model-free reinforcement learning (RL) provides a data-driven solution to the order execution problem. However, the existing works always optimize execution for an individual order, overlooking the practice that multiple orders are specified to execute simultaneously, resulting in suboptimality and bias. In this paper, we first present a multi-agent RL (MARL) method for multi-order execution considering practical constraints. Specifically, we treat every agent as an individual operator to trade one specific order, while keeping communicating with each other and collaborating for maximizing the overall profits. Nevertheless, the existing MARL algorithms often incorporate communication among agents by exchanging only the information of their partial observations, which is inefficient in complicated financial market. To improve collaboration, we then propose a learnable multi-round communication protocol, for the agents communicating the intended actions with each other and refining accordingly. It is optimized through a novel action value attribution method which is provably consistent with the original learning objective yet more efficient. The experiments on the data from two real-world markets have illustrated superior performance with significantly better collaboration effectiveness achieved by our method.
% have provided novel insights on this task and 
\end{abstract}






\maketitle
\section{Introduction}
\begin{figure}[t]%
    \begin{subfigure}[b]{0.47\linewidth}
        \includegraphics[width=\textwidth]{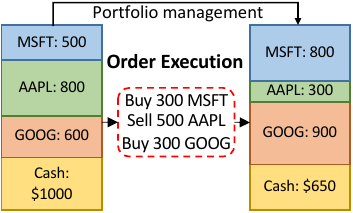}
        \caption{Illustration of financial investment as a combination of portfolio management and multi-order execution.}
        \label{fig:multi-order}%
    \end{subfigure}
    \hfill
    \begin{subfigure}[b]{0.47\linewidth}
        \includegraphics[width=\textwidth]{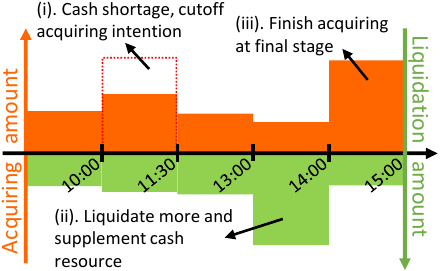}
        \caption{Imbalanced cash utilization requiring further coordination between acquirement and liquidation operations.}
        \label{fig:imbalance}%
    \end{subfigure}
    \vspace{-5pt}
    \caption{An example of multi-order execution and motivation of collaboration within it.
    }
\vspace{-10pt}
\end{figure}

In quantitative finance, the primary goal of the investor is to maximize the long-term value through continuously trading of multiple assets in the market~\cite{wang2019alphastock,ye2020reinforcement}.
The process consists of two parts, portfolio management, which dynamically allocate the portfolio across the assets, and \textit{order execution} whose goal is to fulfill a number of acquisition or liquidation orders specified by the portfolio management strategy, within a time horizon, and close the loop of investment~\cite{cartea2015optimal, fang2021universal}.
Figure~(\ref{fig:multi-order}) presents the trading process within one trading day.
The trader first updates the target portfolio allocation following some portfolio management strategy.
Then, the orders shown in the red dotted zone need to be executed to accomplish the actual portfolio adjustment.
We focus on order execution task in this paper, which aims at \textit{simultaneously finish executing multiple orders} during the time horizon while maximizing the overall execution profit gain.

The challenge of order execution lies in two aspects.
First, the number of orders changes according to the portfolio allocation from day to day, which requires the order execution strategy to be scalable and flexible to support \textit{large and various number of orders}.
Second, \textit{cash balance is limited} and all acquiring operations will consume the limited cash supply of the trader, which can only be replenished by the liquidating operations.
The lack of cash supply may lead to missing good trading opportunities, which urges one to achieve balance between acquisition and liquidation, to avoid conflicted trading decisions that would cause cash shortage and poor trading performance.
Figure~(\ref{fig:imbalance}) illustrates a typical example of a conflicted trading decision that results in cash imbalance and low execution utility.
The execution of acquisition orders are forced to postpone due to cash shortage until the liquidation orders supplementing the cash, leading to missing the best acquisition opportunity.
We have observed similar evidences in real-world transactions and made more analysis in the experiment.

Although there exists many works for order execution, few of them manage to address the above three challenges.
Traditional financial model based methods~\cite{almgren2001optimal,kakade2004competitive,lee2007multiagent} and some recently developed model-free reinforcement learning (RL) methods~\cite{ning2018double,lin2020end,fang2021universal} only optimize the strategy for single-order execution without considering practice of multi-order execution, which would result in low trading efficacy.
Moreover, it is not applicable to directly transfer the existing methods to multi-order execution since utilizing only one agent to conduct the execution of multiple orders would lead to scalability issue 
as the action space of one individual agent grows exponentially with the number of orders. 
Also, it is either not flexible enough for the execution of varying number of orders~\cite{zhou2019factorized,tan1993multi,li2019cooperative}.

To resolve the above challenges, 
we treat multi-order execution as a multi-agent collaboration problem and utilize a multi-agent reinforcement learning (MARL) method where each agent acts to execute one individual order to factorize the joint action space for scalability to varying order numbers~\cite{foerster2018counterfactual,zhang2018fully,wang2021hierarchical}, and all agents collaborate to achieve higher overall profits with less decision conflicts.

However, the existing MARL solutions for general multi-agent collaboration are not suitable for the multi-order execution environment where the actions of one agent can significantly influence the others through shared cash balance, further affecting the final performance as the financial market changes drastically.
The mainstream methods, which build the communication channel among agents to promote collaboration~\cite{das2019tarmac,sukhbaatar2016learning,huang2020learning}, only allow agents to share information of their partial observation, which can not directly reflect the intention of agents, thus harming the collaboration performance.
\cite{kim2020communication} models the intentions of agents as imagined future trajectories and share the intentions through communication, in order to achieve better collaboration performance.
However, it requires predicting environment dynamics and future actions of others to generate the imagined trajectory, which is intractable especially in the noisy and complicated financial market.
Also, the agents herein can not respond to the intentions of others, i.e., changing the actions they intended to take after receiving the messages, until the next timestep, which makes the intention message less helpful for achieving well coordination at the current timestep.

In this paper, we propose a novel multi-round intention-aware communication protocol for communicating the intended actions among agents at each timestep, which is optimized through an action value attribution method.
Specifically, we first model the intention of agents as \textit{the actions they intended to take} at current timestep~\cite{cohen1990intention,tomasello2005understanding} and share the intentions between agents through communication.
Then, during multiple rounds of communication, the agents are allowed to coordinate with each other and achieve better balance between acquisition and liquidation, whereafter the last intended actions are taken as the final decisions of \textit{the current timestep}.
Thus, note that the intended actions of agents 
should be gradually refined for better collaboration during multi-round communication.
To ensure this, we propose a novel action value attribution method to directly optimize and refine the intended actions at each round,
which has been proved as unbiased to the original decision making objective yet more sample efficient.

Our contributions are three-fold as discussed below.
\begin{itemize}[leftmargin=4mm,topsep=2pt]
	\item We illustrate the necessity of \textit{simultaneous} optimization for all orders in multi-order execution task. To the best of our knowledge, this is the first work formulating this problem as a multi-agent collaboration task and utilize MARL method to solve it.
    \item We formulate the intention of agents as the actions they intended to take and propose a novel action value attribution method to optimize the intended actions directly. 
    We are the first to explicitly refine the intended actions of agents in a cooperative scenario, 
    which may shed some light on the researches about general multi-agent reinforcement learning.
    \item Our proposed intention refinement mechanism 
	allows agents to share and modify their intended actions before the final decisions are made. The experiment results on two real-world stock markets have demonstrated the superiority of our approach on both trading performance and collaboration effectiveness.
\end{itemize}

\vspace{-8pt}
\section{Related Work}

\subsection{RL for Order Execution}
Reinforcement learning (RL) based solutions are proposed for order execution due to its nature as a sequential decision making task.
Early works~\cite{hendricks2014reinforcement,hu2016optimal,daberius2019deep} extend traditional control theory based methods~\cite{bertsimas1998optimal,almgren2001optimal} and rely on unrealistic assumptions for the market, thus not performing well in real-world situations.
Several following works~\cite{nevmyvaka2006reinforcement,ning2018double,lin2020end,fang2021universal,7407387} adopt a data-driven mindset and utilized model-free RL methods to learn optimal trading strategies.
However, all these methods target on individual order execution and do not consider practical constraints under multi-order execution as shown in Figure~(\ref{fig:imbalance}), leading to sub-optimal or impractical trading behaviors.
Although MARL has been widely adopted in financial area for market simulation~\cite{bao2019multi,karpe2020multi,lussange2021modelling,lussange2019stock} and portfolio management~\cite{lee2020maps,lee2007multiagent,wang2021commission}, these is no existing method utilizing MARL directly for order execution.
To our best knowledge, this is the first work using MARL for multi-order execution task with practical constraint.

\vspace{-10pt}
\subsection{Communication in General MARL}
Communication is an essential approach to encourage collaboration between agents for multi-agent collaboration problems.
Early works \cite{tan1993multi,melo2011querypomdp,panait2005cooperative} design pre-defined communication protocols between agents.
DIAL~\cite{foerster2016learning} and CommNet~\cite{sukhbaatar2016learning} first proposed differentiable communication mechanism with deep neural networks.
The following works can be divided into two sub-groups.
The first group focuses mainly on ``who to communicate'', i.e., throttling the communication channel rather than a fully connected global one~\cite{jiang2020graph,jiang2018learning,huang2020learning,kim2019learning}.
While the second concentrates on ``how to deliver messages'', i.e., designing different network structures for better information passing and aggregation in communication channels \cite{peng2017multiagent,jiang2020graph,das2019tarmac}.
However, the messages shared in all these communication methods contain only the information from the observations of agents and do not explicitly reflect their intentions, leading to catastrophic discordance.
Our work is orthogonal to these methods since we focus on ``what to communicate'' and the corresponding optimization algorithm, which can be easily adapted to the communication structure in the works mentioned above to make up for their shortcomings.

\subsection{Intention Modeling}
Explicit modeling of the intentions of agents has been used in Theory of Mind (ToM) and Opponent Modeling (OM) methods.
ToM-net \cite{rabinowitz2018machine} captures mental states of other agents and predicts their future action.
OM \cite{raileanu2018modeling} uses agent policies to predict the intended actions of opponents.
But all these works are conducted under a competitive setting and require agents to infer each other's intention, which could be inaccurate considering the instability nature of MARL~\cite{lowe2017multi}.
\cite{kim2020communication} first conducts intention sharing between agents through communication mechanisms under cooperative settings.
However, the agents are not allowed to modify their action after receiving others' intention at the current step, thus still suffering from discordance.
Also, this method requires forecasting the state transitions of the environment of the next few steps, which may suffer from compounding error problems, especially in finance area where the environment is extremely noisy. 
While our method can improve the joint action of agents gradually through the intention sharing and refinement process, and does not use predicted state transitions or any additional environment information.

\vspace{-5pt}
\section{Preliminary}\label{sec:task}
In this section, we first present the task definition of multi-order execution problem, including notations for variables and optimization goals.
Then, we formulate the task as a Markov decision process of the trader interacting with the market environment.

\subsection{Multi-Order Execution}
We generalize the typical settings of order execution in previous works~\cite{cartea2015algorithmic,ning2018double} to \textit{multi-order} execution, where a set of orders need to be fulfilled within a predefined time horizon.
As shown in Figure~(\ref{fig:multi-order}), we take intraday order execution as a running example while the other time horizons follow the same schema.
Thus, for each trading day, there are a set of $n$ orders to be traded for $n$ assets, respectively, which are denoted as
a tuple $( n, c_0, \mathbf{d}, \mathbf{M} )$.
It includes the trading directions of the orders 
$\mathbf{d}=(d^1, \ldots, d^n)$
where $d^i \in \{1, -1\}$ stands for liquidation and acquisition, respectively and order index $i \in [1,n]$;
the amount of shares to trade for each asset 
$\mathbf{M}=(M^1, \ldots, M^n)$;
and the initial cash balance of the trading account $c_0$ which is shared by all the trading operations during the day as explained below.
For simplicity, we assume that there are $T$ timesteps in the time horizon, i.e., one trading day.
At each timestep $t \in [1, T]$,
the trader should propose the volumes to trade, denoted as 
$\mathbf{q}_t = (q_t^1, \ldots, q_t^n)$. 
The market prices of the corresponding assets are
$\mathbf{p}_t = (p_t^1, \ldots, p_t^n)$, 
which would \textit{not} be revealed to the trader before her proposing the trading decision at $t$.
During the trading process, for the trader, liquidating operations replenish the cash balance while acquiring operations consume it.
As a result, the balance of trading account varies every timestep as 
$c_{t} = c_{t-1} + \sum_{i=1}^n d^i\cdot(p^i_{t}\cdot q^i_{t})$.

Different from the previous works for order execution that only optimize for a single order, the objective of multi-order execution is to maximize the overall profits while fulfilling all the orders without running out of cash, which can be formulated as follows:
\begin{equation}
\begin{aligned}
\label{eq:objective}
\argmax_{\mathbf{q}_1, \ldots, \mathbf{q}_{T}}  
        \sum_{i=1}^n
        d^i
        &\left[
        \sum_{t=1}^{T} 
        (p^i_{t}\cdot q^i_{t}) 
        \right] 
        / 
        \left(
        \sum_{t=1}^{T} q^i_{t}
        \right),\\
    \text{s.t.} \sum_{t=1}^{T} \mathbf{q}_{t} =\mathbf{M}, ~&\mathbf{q}_{t} \geq 0, c_t \geq 0, \forall t \in \{1,\ldots,T\}. 
\end{aligned}
\end{equation}

The average execution price (AEP) of order $i$ is calculated as 
\begin{equation}\label{eq:aep}
    \bar{p}^i = 
        \left[
        \sum_{t=1}^{T} 
        (p^i_{t}\cdot q^i_{t}) 
        \right] 
        / 
        \left(
        \sum_{t=1}^{T} q^i_{t}
        \right)~.
\end{equation}
The trader needs to maximize the AEP of all liquidation orders ($d^i=1$) while minimizing that of acquisition orders ($d^i=-1$).

\subsection{Multi-Order Execution as a Markov Decision Process}\label{sec:mdp}
The order execution problem can be formulated as a Markov decision process (MDP) as $(n, \mathcal{S}, \mathcal{A}, R, I, \gamma)$, where each agent executes one order and all the agents share a collective goal. 
Here $\mathcal{S}$ is the state space, $\mathcal{A}$ is the action space for the agent.
$n$ is the number of agents corresponding to the order number.
Note that, for different trading days, order number $n$ varies dynamically and can be large, which makes the joint action space extremely huge for one single-agent RL policy \cite{ning2018double,lin2020end} to learn to execute multiple orders simultaneously.
Thus, in multi-agent RL schema for multi-order execution, we treat each agent as the individual operator for executing one order.
Each agent has a policy $\pi^i(\s^i;\btheta^i)$ which produces the distribution of actions and is parameterized by $\btheta^i$.
The actions of all agents $\bs{a} = (a^1, ..., a^n)$ sampled from each corresponding policy is used to interact with the environment and $R(\s, \bs{a})$ is the reward function.
$I(\s, \bs{a})$ is the transition function of environment which gives the next state 
by receiving the action.
Our goal is to optimize a unified policy $\bs{\pi} = \{\pi^1,...,\pi^n\}$ with parameter $\btheta = \{\btheta^1,...,\btheta^n\}$ to maximize the expected accumulative reward 
\begin{equation}\label{eq:target}
J(\btheta)=E_{\tau\sim p_{\bs{\pi}}(\tau)}\left[\sum_t \gamma^t R(\s_t,\bs{a}_t)\right] ~,
\end{equation}
where $p_{\bs{\pi}}(\tau)$ is the probability distribution of trajectory $\tau = \{(\s_1,\allowbreak \bs{a}_1), \ldots, (\s_T, \bs{a}_T)\}$ sampled by $\bs{\pi}$ from the environment and $\gamma$ is the discount factor.
More implementation details of the MDP definitions are presented below.

\minisection{Number of agents $n$}
We define the agent to be the operator for a single asset, and each agent would be responsible for the execution of the corresponding order.

\minisection{State space $\mathcal{S}$}
The state $\s_t \in \mathcal{S}$ describes the overall information of the system.
However, for each agent executing a specific order $i$, the state $\bs{s}_t^i$ observed at timestep $t$ contains only the historical market information of the corresponding asset collected just \textit{before} timestep $t$ and some shared trading status.
The detailed description of the observed state information has been illustrated in Appendix \ref{app:state}.

\minisection{Action space $\mathcal{A}$}
Agent $i$ proposes the action $a^i_t\in\mathcal{A}$ after observing $\bs{s}_t^i$ at $t \in \{1,\ldots,T\}$.
Following~\cite{fang2021universal}, we define discrete action space as
$a^i_t \in \{0, 0.25, 0.5, 0.75, 1\}$ which corresponds to the proportion of the target order $M^i$.
The trading volume $q^i_{t} = a^i_t M^i$ will be executed at timestep $t$.
Moreover, $\sum_{t=1}^{T}a^i_t=1$ has been satisfied by fixing $a^i_{T} = (1-\sum_{t=1}^{T-1}a^i_t)$ to ensure all orders are fully fulfilled.
The similar setting has been widely adopted in the related literature \cite{ning2018double,lin2020end}.
We also conduct experiments on different action spaces and present the results in Appendix \ref{app:analysis}, which shows that this action setting has  performed sufficiently well.

We should note \textit{cash limitation} here. 
If the remained cash balance is not adequate for all acquisition agents, then the intended trading volume of them are cutoff, by environment, evenly to the scale that the remained cash balance will just be used up after this timestep. 
For instance, if the actions of all the acquisition agents require twice as much as the remained cash, their viable executed volume will be normalized to half, to ensure cash balance $c_t \geq 0$ for $1 \leq t \leq T$.

\minisection{Reward function}
Different from the previous works for order execution that optimize the performance of each order individually, we formulate the reward of all agents in multi-order execution as 
the summation of rewards for execution of each individual order,
which includes three parts: the profitability of trading, the penalty of market impact and the penalty of cash shortage where the cash balance has been used up and the acquisition actions are limited.

First, to account for the profitability during order execution caused by actions, following \cite{fang2021universal}, we formulate this term of reward as volume weighted \textit{execution gain} upon the average price as 
\begin{equation}\label{eq:rew}
    R_{e}^{+}(\s_t, a^i_t;i) = d^i \frac{q^i_{t}}{M^i} \cdot\overbrace{ \left( \frac{p_{t}^i - \Tilde{p}^i }{ \Tilde{p}^i} \right) }^{ \text{price normalization} } 
    = d^i a^i_t \left( \frac{p_{t}^i}{\Tilde{p}^i} - 1 \right), % \cdot d^i ~,
\end{equation}
where $\Tilde{p}^i=\frac{1}{T}\sum_{t=1}^{T}p^i_{t}$ is the average market price of asset $i$ over the whole time horizon.
Note that, as discussed in \cite{fang2021universal}, incorporating $\Tilde{p}^i$ in the reward function at timestep $t$ will not cause information leakage since the reward has not been included in the state $\bs{s}_t$ thus would not influence the actions of our agent. 
It would only take effect in back-propagation during training.

Second, to avoid potential market impacts, i.e., too large trading volume that may harmfully influence the market, we follow the recent works~\cite{ning2018double,lin2020end,fang2021universal} and propose a quadratic penalty for trading too much of asset $i$ within a short time period as
\begin{equation}\label{eq:rew2}
    R_{a}^{-}(\s_t, a^i_t;i)
    = -\alpha (\frac{q^i_t}{M^i})^2 = -\alpha (a^i_t)^2,
\end{equation}
where $\alpha$ is a hyper-parameter controlling the penalty degree.

Third, we also penalize all the agents whenever the remained cash balance is used up right at timestep $t$, as
\begin{equation}\label{eq:rew3}
    R_{c}^{-}(\s_t, a^i_t;i)
    = -\sigma\mathbbm{1}[c_{t} = 0 | c_{t-1} > 0],
\end{equation}
where $\sigma$ is the hyper-parameter. 

Thus, the reward for executing the $i$-th order is defined as
\begin{equation}
\begin{aligned}
\label{eq:self-reward}
    R(\s_t, a^i_t;i) &= R_{e}^{+}(\s_t, a^i_t;i) + R_{a}^{-}(\s_t, a^i_t;i) + R_{c}^{-}(\s_t, a^i_t;i)  \\
    &= d^i a^i_t \left( \frac{p_{t}^i}{\Tilde{p}^i} - 1 \right) -\alpha (a^i_t)^2 - \sigma \mathbbm{1}[c_{t} = 0 | c_{t-1} > 0].
\end{aligned}
\end{equation}

Finally, in order to optimize the execution of all the orders holistically, the overall reward function, which is shared by all the agents, is defined as the average of rewards of all the orders as
\begin{equation}
	R(\s_t, \bs{a}_t) = \frac{1}{n} \sum_{i=1}^n R(\s_t, a^i_t;i) ~.
\end{equation}

\minisection{Assumptions}
There are two main assumptions adopted in this paper.
Similar to \cite{lin2020end},
(i) the temporary market impact has been adopted as a reward penalty, i.e., $R^-_a$, and we assume that the market is resilient and will bounce back to equilibrium at the next timestep.
(ii) We either ignore the commissions and exchange fees as this expense is relatively small fractions for the institutional investors that we mainly aim at.

\section{Methodology}
In this section, we first briefly introduce the main challenges of solving multi-order execution task.
Then, we illustrate how the previous communication based MARL methods fail facing these challenges by describing the design and problems of general framework of existing method, further clarifying the motivation of our improvements.
Finally, we introduce the details of our multi-round intention-aware communication with action refinement, including the optimization method.

\subsection{Problems of General Multi-agent Communication Framework}\label{sec: mac}
There are two main challenges for solving multi-order execution task with MARL method: (1) The agents have to extract essential information from their own observation, e.g., judge whether it is a good opportunity to trade to derive a high profit. (2) The acquisition and liquidation orders should be coordinated with each other to maintain a reasonable cash balance without severe cash shortage.
This requires the agents to be aware of the situation and decision intention of the other agents at the current timestep, and adjust their decisions to avoid potential conflicts, e.g., congested cash consumption.
However, the existing multi-agent communication methods are limited by a common inflexible framework and can not solve all these challenges, as summarized and discussed as below.

Note that, the following procedure is conducted at each timestep, thus, we omit the subscript $t$ of timestep for simplicity in condition that the context is of clarity.

To solve the above challenge (1), at each timestep, the $i$-th agent would extract a hidden state from its observation $\s^i$ with an information extractor $E(\cdot)$
\begin{equation}\label{eq:extractor}
	\bs{h}_0^i = E(\s^i)~.
\end{equation}
Then, the agents would communicate with each other using a communication channel $C(\cdot)$ and update the hidden states, for totally
$K$ rounds 
for a thorough information exchange,
\begin{equation}
	(\bs{h}^1_{k} \ldots, \bs{h}^n_{k}) = C(\bs{h}^1_{k-1}, \ldots, \bs{h}^n_{k-1}), 1\leq k \leq K~.
\end{equation}
Finally, the actions of agents are generated with a decision making module $D(\cdot)$ as
\begin{equation}
	a^i \sim D(\bs{h}_K^i).
\end{equation}

The previous works on multi-agent communication have been summarized above. 
They all focus on improving the structure of communication channels $C$, either throttling the channel for more efficient communication~\cite{jiang2020graph,jiang2018learning,huang2020learning,kim2019learning} or improving the information passing ability with a more complicated channel design~\cite{peng2017multiagent,jiang2020graph,das2019tarmac}.

However, none of these approaches breaks through the above framework, and we claim that two main problems exist within this framework. 
First, all hidden representations $\bs{h}^i_k$ only contain the information of partial observations of the agents but not the actions they intended to take, making it harder for agents to reach good collaboration.
Second, though multiple rounds of communications are conducted during this process,
the agents only make decisions once after the final round of communication. 
They have no opportunity to refine their actions afterward, leading to discordance as it is hard for agents to reach an agreement immediately in a complicated environment like the finance market. 
These problems combined making existing methods fail to solve the challenge (2) mentioned above, thus not suitable for multi-order execution task.

\subsection{Intention-Aware Communication}
In this section, we first describe the framework of our proposed intention-aware communication (IaC) method.
Then, we discuss the optimization details including the intended action refinement.
\subsubsection{Decision making  with multi-round intention communication}\label{sec:mric}

\begin{figure}[t]
    \includegraphics[width=0.9\linewidth]{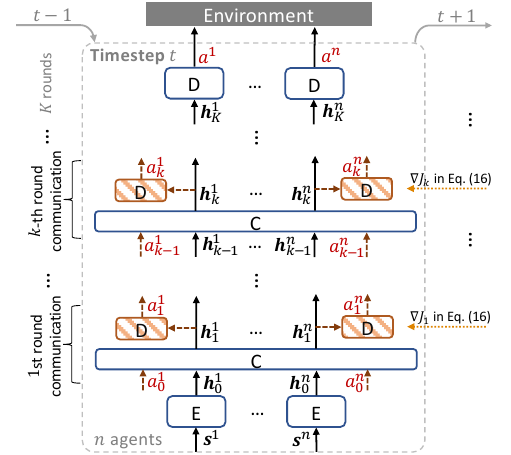}
    \caption{
    The policy framework of our proposed intention-aware communication method. It illustrates the decision-making process of all agents at timestep $t$.
    Totally $K$ rounds of intention-aware communication is conducted among agents to derive the final decision $\bs{a}$.
    Intended actions $\bs{a}_k$ will be made after the $k$-th round and fed to the next communication.
    All the modules of different rounds
    share the same parameters $\btheta$, thus, few more parameters are introduced.
    }
    \label{fig:comm-iac}
    \vspace{-15pt}
\end{figure}

To solve the problems mentioned above, we propose the intention-aware communication, which can be divided into two parts: observation extraction and multi-round communication with decision making.
Following the above convention, we by default omit the subscript $t$ in notations without causing misunderstanding since this procedure is conducted during each timestep.
The whole process has been illustrated in Figure~\ref{fig:comm-iac}.

\minisection{Observable information extraction} 
Similar to the framework described in Section~\ref{sec: mac}, from the view of the $i$-th agent, at each timestep, an information extractor $E$ is utilized to extract the patterns from the input and encode them as an initial hidden representation $\bs{h}_0^i$ of agent $i$ from its observation $\s^i$ following Eq.~(\ref{eq:extractor}).

\minisection{Multi-round communication with decision making}
Our central improvement lands in the multi-round communication process.
Instead of designing more complicated communication channels, we focus on ``what to communicate'' and share the intended actions of agents during each round of communication.
Also, we make it possible for agents to constantly refine their actions according to the intentions of the others during this process.
The process can be formulated as
\begin{align}
        (\bs{h}^1_{k}, \ldots, \bs{h}^n_{k}) &= C(\bs{h}^1_{k-1} || a_{k-1}^i, \ldots, \bs{h}^n_{k-1} || a_{k-1}^n), \label{eq:iac-c}\\ 
        a_k^i &\sim D(\bs{h}^i_k), ~~~~ 1\leq i \leq n, 1\leq k \leq K ~, \label{eq:iac-d}
\end{align}
where $a_0^i$ is a dummy action 
whose value can be arbitrarily assigned, $C(\cdot)$ is the communication channel where agents exchange information and update their hidden states for one round from $\bs{h}_k$ to $\bs{h}_{k+1}$, and $D(\cdot)$ is the decision making module.
The intended actions of the last round are used as the final actions actually executed in the environment as $\bs{a}=\bs{a}_K$ in our method.

Note that, our proposed method is different from the general framework of the previous communication-based MARL methods described in Section~\ref{sec: mac}, which only share the information extracted from the partial observations of agents and make the (final) decisions only once after the last round of communication.
The novel parts of our proposed method are emphasized with dashed lines and striped blocks
in Figure~\ref{fig:comm-iac}.

The communication channel $C(\cdot)$ is scalable to varying number of agents following previous methods~\cite{foerster2016learning,das2019tarmac},

\subsubsection{Intention optimization with action value attribution}
The intended actions $\bs{a}_k$ generated after $k$-th communication round should provide the intention of every agent, which reflects the instant intuition of decision making at the current timestep.
Thus, the intended actions $\bs{a}_k$ should by design reflect the true intentions of agents and be exchanged with each other through the next round of communication to further facilitate collaborative decision making, until the final round as $\bs{a} = \bs{a}_K$.
To achieve this, we propose an auxiliary objective for each round of intended actions.
All auxiliary objectives are optimized together with the original objective $J(\btheta)$ defined in Eq.~(\ref{eq:target}) to keep all intended actions highly correlated to the final goal and progressively refine decisions upon round and round.

We first introduce some definitions before describing the design of the auxiliary objective in detail.
Recall that we set the context of agents at timestep $t$,
we define value function $V(\s_t) = \sum_{t'=t}^T \gamma ^{t'-t} R(\bs{s}_{t'},\bs{a}_{t'})$ as the expected cumulative reward we could get starting from state $\s_t$ following our policy $\bs{\pi}$, and action value $Q(\s_t, \bs{a}_{t}) = R(\s_t, \bs{a}_{t}) + \gamma V(I(\s_t, \bs{a}_{t}))$ as the expected cumulative reward if we take action $\bs{a}_{t}$ at timestep $t$ following policy $\bs{\pi}$.

Denoting the intention generation process during the $k$-th round of communication as $\bs{\pi}_k$, i.e., $\bs{\pi}_k(\cdot | \s, \bs{a}_{k-1})=D(C(\bs{h}_{k-1}||\bs{a}_{k-1})),$ we optimize $\bs{\pi}_k$ by defining an auxiliary objective function to maximize the expected cumulative reward $Q(\s, \bs{a}_{k})$ \textit{as if} we took $\bs{a}_k$ as the final actions instead of $\bs{a}_K$, for all $\s$ and $\bs{a}_{k-1}$ encountered during interacting with the environment.
Note that, when $k = K$, this objective is consistent with our original objective $J(\btheta)$ as $\bs{a}_K$ are the final actions that are used to interact with the environment.
Thus, all our objectives can be uniformly denoted as
\begin{equation}\label{eq:aux-object}
    J_k(\btheta) = \mathbb{E}_{\bs{s},\bs{a}_{k-1}}\left[\mathbb{E}_{\bs{a}_k \sim \bs{\pi}_k(\cdot|\s, \bs{a}_{k-1};\btheta)}[Q(\s, \bs{a}_k)]\right], 1 \leq k \leq K ~,
\end{equation}
where $\btheta$ are the parameters of $E(\cdot)$, $C(\cdot)$, $D(\cdot)$.
The gradient of $J_k$ w.r.t. $\btheta$ is calculated as
\begin{equation}\label{eq:pg}
\nabla_{\btheta} J_k(\btheta) = \mathbb{E}_{\bs{s}, \bs{a}_k, \bs{a}_{k-1}}[\nabla_{\btheta}\log\bs{\pi}_k(\bs{a}_k | \bs{s}, \bs{a}_{k-1};\btheta) Q(\bs{s}, \bs{a}_k)] ~.
\end{equation}

We further expect the intended actions should generally be refined during the communication process, which means $\bs{a}_k$ should achieve higher return than $\bs{a}_{k-1}$.
Therefore, we use the expected return of the intended actions in last round of communication as a baseline function into Eq.~(\ref{eq:pg}) where $Q(\bs{s}, \bs{a}_0)=0$,
\begin{multline}\label{eq:adv-attri}
\nabla_{\btheta} J_k(\btheta) = \mathop{\mathbb{E}}_{\bs{s}, \bs{a}_k, \bs{a}_{k-1}}[\nabla_{\btheta}\log\bs{\pi}_k(\bs{a}_k | \bs{s}, \bs{a}_{k-1};\btheta) \\ (Q(\bs{s}, \bs{a}_k) - Q(\bs{s}, \bs{a}_{k-1}))] ~.
\end{multline}
It is reasonable as we would like to encourage $\bs{a}_{k}$ to achieve better performance than $\bs{a}_{k-1}$ and penalize those perform worse.
Moreover, the policy gradient in Eq.~(\ref{eq:adv-attri}) remains unbiased to Eq.~(\ref{eq:pg}) since the gradient of the baseline w.r.t. $\btheta$ is
\begin{equation}
\begin{aligned}
\label{eq:unbiased}
    &\mathbb{E}_{\bs{s}, \bs{a}_k, \bs{a}_{k-1}}[-\nabla_{\btheta}\log\bs{\pi}_k(\bs{a}_k | \bs{s}, \bs{a}_{k-1} ;  \btheta) Q(\bs{s}, \bs{a}_{k-1})]  \\
    =& \mathbb{E}_{\bs{s},\bs{a}_{k-1}}[-Q(\bs{s}, \bs{a}_{k-1}) \mathbb{E}_{\bs{a}_k}[\nabla_{\btheta}\log\bs{\pi}_k(\bs{a}_k | \bs{s} , \bs{a}_{k-1};  \btheta)]]  \\
    =& \mathbb{E}_{\bs{s},\bs{a}_{k-1}}[-Q(\bs{s}, \bs{a}_{k-1}) \nabla_{\btheta}1] = 0 ~.
\end{aligned}
\end{equation}
\minisection{Action value attribution}
We further clarify the intuition of our auxiliary objective from a perspective of attributing the credit of final decision making across the whole process of intention refinement during each round of communication.
First, aiming at encouraging the agents to find better actions than the intended actions exchanged during last round of communication, we define another auxiliary objective for optimizing $\bs{\pi}_k$ as
\begin{equation}\label{eq:delta-q}
J'_k(\btheta) = \mathbb{E}_{\bs{s},\bs{a}_{k-1}}\left[\mathbb{E}_{\bs{a}_k \sim \bs{\pi}_k(\cdot|\s, \bs{a}_{k-1};\btheta)}[Q(\s, \bs{a}_k) - Q(\s, \bs{a}_{k-1})]\right].
\end{equation}
Taking derivatives of Eq.~(\ref{eq:delta-q}) w.r.t. $\btheta$ and considering Eq.~(\ref{eq:adv-attri}), we can easily find that $\nabla_{\btheta} J_k(\btheta) = \nabla_{\btheta} J'_k(\btheta)$.
Thus, considering the consistency between $J_K(\btheta)$ and $J(\btheta)$ we mentioned above, $J'_K(\btheta)$, i.e., the auxiliary objective defined over the decisions of the last round, is also consistent with our original target $J(\btheta)$.
Also, as $\sum_{k=1}^K(Q(\s, \bs{a}_k) - Q(\s, \bs{a}_{k-1})) = Q(\s, \bs{a}_K)$, we can see that 
the original optimization goal $J(\btheta)$, i.e., $J'_K$ here,
has been distributed to each $J'_k$ for $1\leq k \leq K$.
We can tell that what we are doing is designing an \textit{action value attribution} method where the value of the last decision $\bs{a}_K$ is attributed to all the intended actions.

Optimization with action value attribution decomposes the final optimization objective to each round of intention-aware communication, which not only alleviates the burden of multi-agent communication optimization, but also improves decision making gradually through learning to promote action value at each round as shown in Eqs.~(\ref{eq:adv-attri}) and (\ref{eq:delta-q}).

\minisection{Action value estimation}
The last detail is to estimate the action value $Q(\s, \bs{a})$ to calculate $\nabla_{\btheta} J_k(\btheta)$ in Eq.~(\ref{eq:adv-attri}).
Normally, for $k=K$, $Q(\s, \bs{a}_K)$ can be directly calculated from the original sampled trajectory as $\bs{a}_K$ is the final decision utilized to interact with the environment.
While, for $1 \leq k < K$, we train an action value estimation model $\hat{Q}(\s, \bs{a})$ upon the trajectories collected by interacting with the environment using the actual decision $\bs{a}_K$.
Note that, this procedure does not require the environment to provide any additional information about the intended actions, which guarantees generalizability for wider applications of MARL.

Overall speaking, we optimize the objective functions $J_k(\btheta)$ for all communication rounds simultaneously, thus, the final loss function to minimize w.r.t. the parameter $\btheta$ is defined as
\begin{equation}\label{eq:overall-loss}
    L(\btheta) = -\frac{1}{K} \sum_{k=1}^K J_k(\btheta) ~.
\end{equation}
As for implementation, we use PPO algorithm \cite{schulman2017proximal} to optimize all the intended actions and the final decisions.

The overall decision-making and optimization process of our proposed Intention-Aware Communication method is presented in Algorithm~\ref{alg:iac}.
The detailed network structures of the extractor $E(\cdot)$, communication channel $C(\cdot)$, decision module $D(\cdot)$ and action value estimator $\hat{Q}(\cdot, \cdot)$ are presented in Section~\ref{app:network}.

\vspace{-5pt}
\begin{algorithm}
\caption{Intention-Aware Communication}\label{alg:iac}
\begin{algorithmic}
\Require Random initialized network parameters $\btheta$
\While{$\btheta$ not converged}
	\State Sample a set of orders $( n, c_0, \mathbf{d}, \mathbf{M} )$ from the dataset and obtain the initial state $\s_1$.
	\For{timestep $t \in [1, T]$}
		\State $\bs{h}^i_{t,0} \gets E(\s^i_t)$ for agent $i \in [1,n]$.
		\State  $a_{t,0}^i \gets $ dummy action for agent $i \in [1,n]$.
		\For{communication round $k \in [1,K]$}
			\State Refine the intended actions $\bs{a}_{t,k}$ as in Eq. (\ref{eq:iac-c}, \ref{eq:iac-d}).
		\EndFor
		\State Obtain $\s_{t+1} = I(\s_t, \bs{a}_{t, K})$ from the environment.
	\EndFor
	\State Update $\btheta$ minimizing $L(\btheta)$ defined in Eq.~(\ref{eq:overall-loss}).\EndWhile
\end{algorithmic}
\end{algorithm}
\vspace{-5pt}

\section{Experiments}
In this section, we present the experiment settings and results with extended investigations. 
The reproducible codes and benchmark with data will be released upon the acceptance of this paper.

\subsection{Datasets}
All the compared methods are trained and evaluated based on the historical transaction data of China A-share stock market US stock market from 2018 to 2021 collected from Yahoo Finance\footnote{\url{https://finance.yahoo.com}}.
All datasets are divided into training, validation, and test sets according to time and the statistics of all datasets are presented in Appendix~\ref{app:dataset}
For both stock markets, we conduct three datasets with a rolling window of one-year length and stride size, denoted as CHW1, CHW2, CHW3 and USW1, USW2, USW3.
For each trading day, several sets of orders with the corresponding initial cash budgets $\{(n, c_0, \mathbf{d}, \mathbf{M})\}$ are generated according to a widely used portfolio management strategy ``Buying-Winners-and-Selling-Losers''~\cite{wang2019alphastock,jegadeesh1993returns} implemented in \cite{yang2020qlib}, and each set of intraday execution orders includes the information about the asset name, the amount and trading type of each order, as discussed in Sec.~\ref{sec:task}.
The orders are the same for all the compared methods for fairness.
Without loss of generality, all the orders in our datasets are restricted to be fulfilled within a trading day, which is 240-minute length for Chinese stock market and 390-minute for US stock market. The detailed trading process has been described in Appendix~\ref{app:trading}.

\subsection{Evaluation Settings}
\subsubsection{Compared methods}
Since we are the first to study the simultaneous optimization for multi-order execution, we first compare our proposed method and its variants with traditional financial model based methods and some single-agent RL baselines proposed for order execution problem.
Note that, for single-agent RL methods that are optimized for single-order optimization, instead of using the summation of rewards for all orders as the reward function, the agent is only optimized for the reward of each individual order $R(\s_t, a_t^i; i)$ as defined in Eq.~(\ref{eq:self-reward}), aligned with how these methods are originally proposed.
Then, to illustrate the effectiveness of our proposed novel intention-aware communication mechanism, we compare our method with those general RL works incorporating common MARL algorithms.
All methods are evaluated with the same multi-order execution procedure described in Appendix~\ref{app:trading}.
\begin{itemize}[leftmargin=9pt]
    \item \textbf{TWAP} (Time-Weighted Average Price) \cite{bertsimas1998optimal} is a passive rule-based method which evenly distributes the order amount across the whole time horizon, whose average execution price would be the average market price $\Tilde{p}^i$ as defined in Eq.~(\ref{eq:rew}).
    \item \textbf{VWAP} (Volume-Weighted Average Price) \cite{kakade2004competitive} is another widely used strategy which distributes the order proportionally to the estimated market volume to reduce market impacts.
    \item \textbf{AC} is derived in \cite{almgren2001optimal} as a trading strategy focusing on balancing price risk and market impacts using close-form market heuristics.
    \item \textbf{DDQN} (Double Deep Q-network)~\cite{ning2018double} is a single-agent value-based RL method for single-order execution optimization.
    \item \textbf{PPO} was proposed in \cite{lin2020end} which utilizes PPO algorithm to optimize single-order execution.
    \item \textbf{CommNet} \cite{foerster2016learning} first utilizes a neural network as a broadcasting communication channel to share information between agents.
    \item \textbf{TarMAC} \cite{das2019tarmac} is a multi-agent reinforcement learning method which first utilizes an attention-based communication channel.
    \item \textbf{IS} \cite{kim2020communication} incorporates intention communication in MARL, which forecasts the future trajectories of other agents as intentions.
    \item \textbf{IaC} is our proposed method, which utilizes an intention-aware communication mechanism to increase the cooperative trading efficacy for multi-order execution. Specifically, for comprehensive comparison, we conduct experiments on two variants of our method, \textbf{\IACT} and \textbf{\IACC}, which utilize the implementation of communication channel as TarMAC and CommNet,
    respectively. 
\end{itemize}
For all the compared methods, the hyper-parameters are tuned on the validation sets and then evaluated on the test sets.
For RL-based methods, the policies are trained with six different random seeds after determining the optimal hyper-parameters and the means and standard deviations of results on test sets are reported.
The detailed hyper-parameter settings are presented in Appendix~\ref{app:para}.
All RL-based methods share the same network structures for extractor $E(\cdot)$, communication network $C(\cdot)$ (if exists) and decision module $D(\cdot)$, thus the sizes of parameters are similar, for fair comparison.

\subsubsection{Evaluation metrics}
The first evaluation metric is the average execution gain (\textbf{EG}) 
over all orders which is calculated as
$\text{EG}=\frac{1}{|\mathbb{D}|}\sum_{i=1}^{|\mathbb{D}|} \text{EG}^i$, where $|\mathbb{D}|$ is the number of orders in the dataset, and $\text{EG}^i$ is the execution gain on order $i$ relative to the corresponding daily average market price $\Tilde{p}^i$ of asset $i$ and defined as $\text{EG}^i = d^i(\frac{\bar{p}^i_\text{strategy} - \Tilde{p}^i}{\Tilde{p}^i}) \times 10^4 \text{\textpertenthousand} $.
Here $\bar{p}_{\text{strategy}}^i$ is the average execution price of the evaluated strategy as defined in Eq.~(\ref{eq:aep}).
Note that EG is proportional to the reward $R_{e}^{+}$ described in Eq.~(\ref{eq:rew}) and has been widely used in order execution task \cite{ning2018double,lin2020end,fang2021universal}.
EG is measured in basis points (BPs), where one basis point is 1{\textpertenthousand}.
To better illustrate the profit ability, we also report the additional annualized rate of return (\textbf{ARR}) brought by the order execution algorithm, relative to the same portfolio management strategy with TWAP execution solution whose average execution price is $\Tilde{p}^i$ and EG$=0$, whose detailed calculation is presented in Appendix~\ref{app:arr}.
Following \cite{ning2018double, lin2020end}, we also report the gain-loss ratio (\textbf{GLR}) of EG $\frac{\mathbb{E}_i[\text{EG}_i | \text{EG}_i > 0]}{\mathbb{E}_i[-\text{EG}_i | \text{EG}_i < 0]}$ and positive rate (\textbf{POS}) $\mathbb{P}[\text{EG}_i > 0]$ across all the orders in the dataset.
All the above metrics, i.e., EG, ARR, GLR and POS, are better while value getting higher.

Moreover, as a reasonable execution strategy should manage the cash resources wisely to avoid shortages, our last evaluation metric is the average percentage of time of conflict (\textbf{TOC}) where the agents conduct conflicted actions and suffer from short of cash during the execution, defined as $100\% \times \mathbb{E}[\sum_{t=1}^{T}\mathbbm{1}(c_t = 0)] / T~.$
TOC is better when the value is lower.
Generally, a high TOC value indicates that the acquisition orders are often limited by the cash supply, which would usually results in suboptimal EG results.
Note that, although the portfolio management strategy responsible to generate daily orders can hold more cash, i.e., allocating a larger initial cash balance $c_0$ for each set of orders to offer more budgets for acquisition and reduce TOC, a large cash position would lower the capital utilization and cause a lower profit rate.
Thus, the initial cash budget $c_0$ would not be very large which requires multi-order execution to actively coordinates among liquidation and acquiring to well manage cash resources, i.e., achieving low TOC value.

\begin{table*}[h]
\resizebox{\textwidth}{!}{\large
\begin{tabular}{|c|c|rrrrr|rrrrr|rrrrr|}
\hline
\multirow{2}{*}{Method Group} & \multirow{2}{*}{Method} & \multicolumn{5}{c|}{China A-share Market Window 1 (CHW1)} & \multicolumn{5}{c|}{China A-share Market Window 2 (CHW2)} & \multicolumn{5}{c|}{China A-share Market Window 3 (CHW3)} \\ \cline{3-17} 
 &  & \multicolumn{1}{c}{EG (\textpertenthousand) $\uparrow$} & \multicolumn{1}{c}{ARR (\%) $\uparrow$} & \multicolumn{1}{c}{POS $\uparrow$} & \multicolumn{1}{c}{GLR $\uparrow$} & \multicolumn{1}{c|}{TOC (\%) $\downarrow$} & \multicolumn{1}{c}{EG (\textpertenthousand) $\uparrow$} & \multicolumn{1}{c}{ARR (\%) $\uparrow$} & \multicolumn{1}{c}{POS $\uparrow$} & \multicolumn{1}{c}{GLR $\uparrow$} & \multicolumn{1}{c|}{TOC (\%) $\downarrow$} & \multicolumn{1}{c}{EG (\textpertenthousand) $\uparrow$} & \multicolumn{1}{c}{ARR (\%) $\uparrow$} & \multicolumn{1}{c}{POS $\uparrow$} & \multicolumn{1}{c}{GLR $\uparrow$} & \multicolumn{1}{c|}{TOC (\%) $\downarrow$} \\ \hline
\multirow{3}{*}{\begin{tabular}[c]{@{}c@{}}Financial model based\\ (single-order optimization)\end{tabular}} & TWAP & 0.00 & 0.00 & 0.50 & 1.00 & 0.00 & 0.00 & 0.00 & 0.50 & 1.00 & 0.00 & 0.00 & 0.00 & 0.50 & 1.00 & 0.00 \\
 & AC & -3.26 & -0.65 & 0.48 & 1.00 & 0.00 & -1.25 & -0.25 & 0.49 & 0.96 & 0.00 & -6.14 & -1.22 & 0.48 & 0.95 & 0.00 \\
 & VWAP & -3.26 & -0.65 & 0.49 & 1.01 & 0.00 & -2.23 & -0.45 & 0.48 & 0.92 & 0.00 & -6.13 & -1.22 & 0.48 & 0.95 & 0.00 \\ \hline
\multirow{2}{*}{\begin{tabular}[c]{@{}c@{}}Single-agent RL\\ (single-order optimization)\end{tabular}} & PPO & 21.63$\pm$1.45 & 5.56$\pm$0.36 & 0.59$\pm$0.01 & 1.27$\pm$0.05 & 40.39$\pm$6.66 & 24.36$\pm$2.32 & 6.28$\pm$0.58 & 0.60$\pm$0.01 & 1.23$\pm$0.04 & 45.12$\pm$10.95 & 20.01 $\pm$ 1.11 & 5.13$\pm$0.28 & 0.58$\pm$0.01 & 1.26$\pm$0.03 & 30.24$\pm$5.37 \\
 & DDQN & 6.25$\pm$0.27 & 1.57$\pm$0.07 & 0.53$\pm$0.01 & 1.05$\pm$0.01 & 8.43$\pm$1.72 & 7.12$\pm$0.64 & 1.80$\pm$0.16 & 0.54$\pm$0.01 & 1.06$\pm$0.03 & 18.27$\pm$3.23 & 7.07$\pm$0.56 & 1.78$\pm$0.14 & 0.53$\pm$0.02 & 1.03$\pm$0.01 & 10.27$\pm$1.14 \\ \hline
\multirow{5}{*}{\begin{tabular}[c]{@{}c@{}}Multi-agent RL\\ (multi-order optimization)\end{tabular}} & CommNet & 20.32$\pm$0.98 & 5.21$\pm$0.25 & 0.60$\pm$0.01 & 1.20$\pm$0.01 & 2.45$\pm$0.43 & 30.21$\pm$1.89 & 7.84$\pm$0.47 & 0.59$\pm$0.02 & 1.33$\pm$0.03 & 6.87$\pm$1.29 & 21.02$\pm$1.24 & 5.39$\pm$0.31 & 0.58$\pm$0.01 & 1.30$\pm$0.01 & 2.98$\pm$0.99 \\
 & TarMAC & 22.46$\pm$1.42 & 5.77$\pm$0.36 & 0.57$\pm$0.03 & 1.29$\pm$0.01 & 2.56$\pm$0.10 & 31.12$\pm$0.88 & 8.09$\pm$0.22 & 0.60$\pm$0.01 & 1.38$\pm$0.02 & 3.75$\pm$0.48 & 21.89$\pm$0.70 & 5.62$\pm$0.18 & 0.58$\pm$0.01 & 1.33$\pm$0.01 & 4.03$\pm$1.53 \\
 & IS & 21.22$\pm$2.88 & 5.45$\pm$0.72 & 0.58$\pm$0.01 & 1.28$\pm$0.02 & 3.02$\pm$0.52 & 30.01$\pm$0.35 & 7.79$\pm$0.09 & 0.59$\pm$0.01 & 1.35$\pm$0.04 & 5.23$\pm$0.51 & 22.04$\pm$1.13 & 5.66$\pm$0.28 & 0.59$\pm$0.01 & 1.37$\pm$0.05 & 4.54$\pm$0.56 \\
 & \IACC & \textbf{28.38$\pm$2.34*} & \textbf{7.35$\pm$0.59*} & \textbf{0.64$\pm$0.02*} & 1.33$\pm$0.01 & 1.56$\pm$0.05 & 32.28$\pm$0.21 & 8.40$\pm$0.05 & 0.60$\pm$0.01 & 1.38$\pm$0.01 & 1.63$\pm$0.33 & 24.78$\pm$1.02 & 6.39$\pm$0.26 & 0.60$\pm$0.01 & \textbf{1.39$\pm$0.01*} & \textbf{1.29$\pm$0.43*} \\
 & \IACT & 28.36$\pm$3.45 & 7.35$\pm$0.86 & 0.63$\pm$0.03 & \textbf{1.34$\pm$0.01*} & \textbf{1.23$\pm$0.36*} & \textbf{33.01$\pm$0.18*} & \textbf{8.60$\pm$0.05*} & \textbf{0.61$\pm$0.01*} & \textbf{1.41$\pm$0.01*} & \textbf{1.58$\pm$0.32*} & \textbf{25.45$\pm$1.22*} & \textbf{6.57$\pm$0.31*} & \textbf{0.60$\pm$0.00*} & 1.38$\pm$0.02 & 1.38$\pm$0.48 \\ \hline
\end{tabular}}
\end{table*}

\begin{table*}[h]
\resizebox{\textwidth}{!}{
\begin{tabular}{|c|c|rrrrr|rrrrr|rrrrr|}
\hline
\multirow{2}{*}{Method Group} & \multirow{2}{*}{Method} & \multicolumn{5}{c|}{US Stock Market Window 1 (USW1)} & \multicolumn{5}{c|}{US Stock Market Window 2 (USW2)} & \multicolumn{5}{c|}{US Stock Market Window 3 (USW3)} \\ \cline{3-17} 
 &  & \multicolumn{1}{c}{EG (\textpertenthousand) $\uparrow$} & \multicolumn{1}{c}{ARR (\%) $\uparrow$} & \multicolumn{1}{c}{POS $\uparrow$} & \multicolumn{1}{c}{GLR $\uparrow$} & \multicolumn{1}{c|}{TOC (\%) $\downarrow$} & \multicolumn{1}{c}{EG (\textpertenthousand) $\uparrow$} & \multicolumn{1}{c}{ARR (\%) $\uparrow$} & \multicolumn{1}{c}{POS $\uparrow$} & \multicolumn{1}{c}{GLR $\uparrow$} & \multicolumn{1}{c|}{TOC (\%) $\downarrow$} & \multicolumn{1}{c}{EG (\textpertenthousand) $\uparrow$} & \multicolumn{1}{c}{ARR (\%) $\uparrow$} & \multicolumn{1}{c}{POS $\uparrow$} & \multicolumn{1}{c}{GLR $\uparrow$} & \multicolumn{1}{c|}{TOC (\%) $\downarrow$} \\ \hline
\multirow{3}{*}{\begin{tabular}[c]{@{}c@{}}Financial model based\\ (single-order optimization)\end{tabular}} & TWAP & 0.00 & 0.00 & 0.50 & 1.00 & 0.00 & 0.00 & 0.00 & 0.50 & 1.00 & 0.00 & 0.00 & 0.00 & 0.50 & 1.00 & 0.00 \\
 & AC & 1.23 & 0.24 & 0.51 & 1.01 & 0.00 & 0.75 & 0.05 & 0.50 & 1.01 & 0.00 & 1.52 & 0.29 & 0.51 & 1.01 & 0.00 \\
 & VWAP & -2.34 & -0.60 & 0.49 & 0.96 & 0.00 & -2.88 & -0.89 & 0.48 & 0.98 & 0.00 & -1.25 & -0.24 & 0.49 & 0.99 & 0.00 \\ \hline
\multirow{2}{*}{\begin{tabular}[c]{@{}c@{}}Single-agent RL\\ (single-order optimization)\end{tabular}} & PPO & 4.02$\pm$0.46 & 1.01$\pm$0.12 & 0.53$\pm$0.01 & 1.03$\pm$0.02 & 14.18$\pm$3.43 & 4.39$\pm$0.84 & 1.10$\pm$0.21 & 0.53$\pm$0.01 & 1.04$\pm$0.03 & 12.36$\pm$2.73 & 4.59$\pm$0.78 & 1.15$\pm$0.20 & 0.54$\pm$0.01 & 0.99$\pm$0.04 & 10.18$\pm$2.53 \\
 & DDQN & 1.08$\pm$0.36 & 0.27$\pm$0.09 & 0.51$\pm$0.01 & 1.02$\pm$0.01 & 7.12$\pm$1.04 & 2.09$\pm$0.77 & 0.52$\pm$0.19 & 0.51$\pm$0.01 & 1.02$\pm$0.01 & 7.12$\pm$1.04 & 1.79$\pm$0.45 & 0.45$\pm$0.11 & 0.52$\pm$0.01 & 0.98$\pm$0.05 & 6.34$\pm$1.44 \\ \hline
\multirow{5}{*}{\begin{tabular}[c]{@{}c@{}}Multi-agent RL\\ (multi-order optimization)\end{tabular}} & CommNet & 6.01$\pm$1.31 & 1.51$\pm$0.33 & 0.53$\pm$0.01 & 1.04$\pm$0.02 & 1.33$\pm$0.53 & 5.75$\pm$1.03 & 1.45$\pm$0.26 & 0.52$\pm$0.01 & 1.04$\pm$0.01 & 1.03$\pm$0.38 & 6.42$\pm$1.83 & 1.62$\pm$0.46 & 0.54$\pm$0.01 & 1.04$\pm$0.01 & 1.88$\pm$0.47 \\
 & TarMAC & 6.54$\pm$1.05 & 1.65$\pm$0.27 & 0.54$\pm$0.01 & 1.03$\pm$0.01 & 2.34$\pm$0.68 & 6.28$\pm$1.52 & 1.58$\pm$0.39 & 0.53$\pm$0.01 & 1.04$\pm$0.02 & 2.58$\pm$0.37 & 6.99$\pm$2.01 & 1.76$\pm$0.51 & 0.54$\pm$0.02 & 1.04$\pm$0.01 & 3.21$\pm$0.88 \\
 & IS & 5.77$\pm$0.88 & 1.45$\pm$0.22 & 0.53$\pm$0.01 & 1.03$\pm$0.01 & 1.67$\pm$0.78 & 4.98$\pm$0.75 & 1.25$\pm$0.19 & 0.52$\pm$0.02 & 1.03$\pm$0.01 & 3.23$\pm$0.58 & 5.85$\pm$0.78 & 1.47$\pm$0.20 & 0.53$\pm$0.01 & 1.03$\pm$0.02 & 4.33$\pm$1.62 \\
 & \IACC & 7.82$\pm$0.25 & 1.97$\pm$0.06 & 0.54$\pm$0.01 & \textbf{1.07$\pm$0.03*} & 1.30$\pm$0.10 & 7.57$\pm$0.32 & 1.91$\pm$0.08 & \textbf{0.55$\pm$0.02*} & 1.06$\pm$0.02 & \textbf{0.97$\pm$0.08*} & 8.01$\pm$0.33 & 2.02$\pm$0.08 & 0.54$\pm$0.01 & \textbf{1.08$\pm$0.03*} & \textbf{1.01$\pm$0.13*} \\
 & \IACT & \textbf{8.11$\pm$0.52*} & \textbf{2.05$\pm$0.13*} & \textbf{0.55$\pm$0.02} & 1.06$\pm$0.01 & \textbf{1.01$\pm$0.22*} & \textbf{7.99$\pm$0.29*} & \textbf{2.02$\pm$0.07} & 0.54$\pm$0.01 & \textbf{1.07$\pm$0.03*} & 1.01$\pm$0.29 & \textbf{8.23$\pm$0.56*} & \textbf{2.08$\pm$0.14*} & \textbf{0.55$\pm$0.02*} & 1.07$\pm$0.01 & 1.38$\pm$0.27 \\ \hline
\end{tabular}
}

\caption{The results of all the compared methods on five rolling windows of two real-world markets. $\uparrow$ ($\downarrow$) means the higher (lower) value is better. For learning-based methods, we report the mean value $\pm$ standard deviation over six random seeds. The best results of learning-based methods are highlighted with bold format. * indicates p-value $<10^{-6}$ in significance test~\cite{bhattacharya2002median}.
}
\label{tab:result-table}\vspace{-20pt}
\end{table*}

\vspace{-5pt}
\subsection{Experiment Results}
The detailed experiment results are listed in Table~\ref{tab:result-table}.
We can tell that 
(1) our proposed methods, i.e., \IACC and \IACT, significantly improve the trading performance on the test environment compared to the other baselines and achieve the highest profits, i.e., EG, POS and GLR on all datasets.
Also, the intention-aware communication mechanism brings a significant reduction in the TOC metric, which proves that sharing intended actions through communication offers much better collaboration performance than the previous MARL methods.
(2) Almost all the MARL methods with multi-order optimization achieve higher profits and lower TOC than the RL methods optimized for single-order.
It has illustrated the necessity of jointly optimizing multi-order execution and encouraging the collaboration among agents.
(3) Although IS also shares the intentions of agents through communication, it achieves worse results than \IAC, indicating that the refinement of intended actions for multiple rounds within a single timestep is important for agents to reach good collaboration in a complicated environment.
Also, intention communication in IS requires predicting future trajectories of all agents, which might not accurately reflect the true intention of other agents.
Moreover, it suffers from large compounding error in noisy financial environments.
(4) All the financial model based methods achieve TOC equal to zero since they do not actively seek best execution opportunity but mainly focus on reducing market impacts~\cite{almgren2001optimal}, which may easily derive low TOC value yet low (poorer) EG performance. 
(5) There exists a huge gap between the performances of RL-based methods on China A-share market and US stock market.
The reason may be the relatively larger daily price volatility in China stock market as shown in
Appendix~\ref{app:market}.

\vspace{-8pt}
\subsection{Extended Investigation}\label{sec:ext-invest}
To further present the necessity of jointly optimizing multiple orders and the improvement in collaboration efficiency achieved by the proposed intention-aware communication mechanism, we investigate the statistics of the market data and compare the transaction details of our \IAC strategy and other baselines. 
The analysis in this section is based on the trading results on the test set of CHW1 and USW1 while the other datasets share similar conclusions.

\begin{figure*}[t]%
    \centering
    \begin{subfigure}{0.6\columnwidth}
        \centering
        \includegraphics[width=\textwidth]{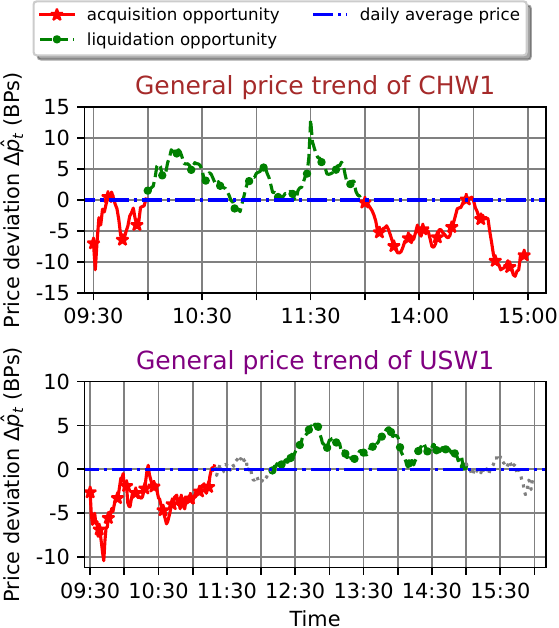}
        \caption{$\mathbbm{E}_t[\Delta\hat{p}_t]$, average price deviation at every minute $\Delta\hat{p}_t = (\frac{p_t}{\Tilde{p}}-1)$ to the averaged price $\Tilde{p}$ over all the market stocks.
        }
        \label{fig:price-trend}%
    \end{subfigure}
    \hfill
    \begin{subfigure}{0.8\columnwidth}
        \centering
        \includegraphics[width=\textwidth]{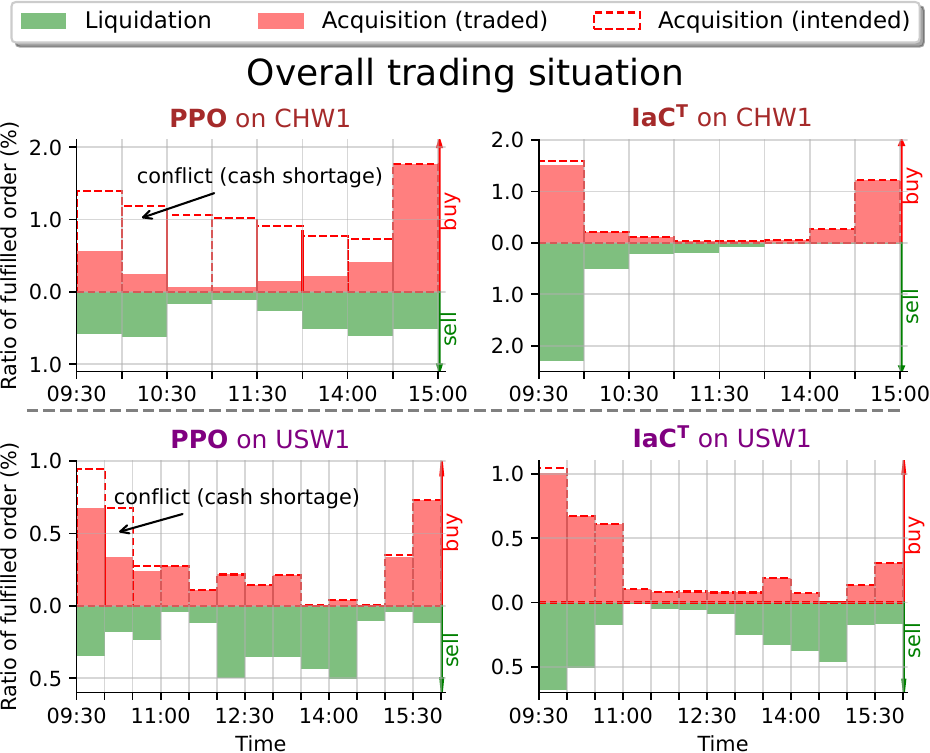}
        \caption{The average ratio of order fulfilled at every minute.
        Better coordination of our \IAC methods leads to higher profits (EG) and lower TOC (i.e., fewer conflicts).
        }
        \label{fig:overall-transaction}
    \end{subfigure}
    \hfill
    \begin{subfigure}{0.5\columnwidth}
        \centering
        \includegraphics[width=\textwidth]{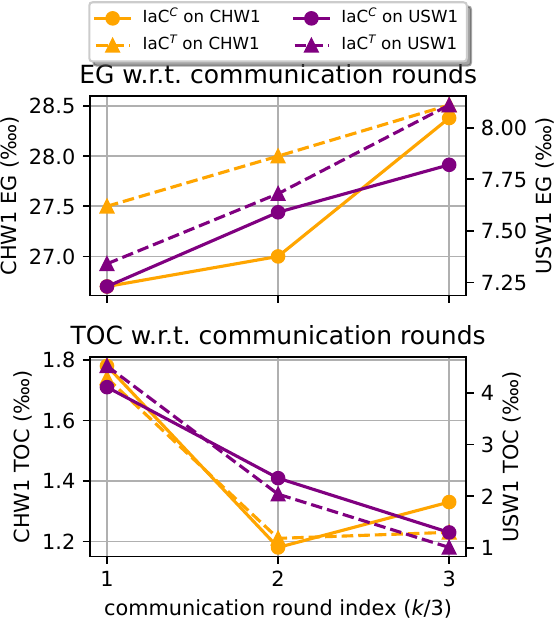}%
        \caption{The TOC and EG of intended actions after different communication rounds in \IACC and \IACT (totally $K=3$ rounds).
        }
        \label{fig:diff-comm}
    \end{subfigure}
    \caption{The price trend of assets in the order and the corresponding trading situation of compared methods.}
\end{figure*}

\minisection{Collaboration is necessary when conducting multi-order execution}
We further clarify the necessity of collaboration for multi-order execution by illustrating the trading opportunity of acquisition and liquidation orders over the trading day.
Figure~\ref{fig:price-trend} illustrates the general price trend of all the assets by showing the average of price deviation at each minute $\Delta \hat{p}_t = \mathbbm{E}_t[\frac{p_t - \Tilde{p}}{\Tilde{p}}]$ as defined in Eq.~(\ref{eq:rew}).
It shows that, on average, there exists acquisition opportunity (lower price) at the beginning of the trading day, while the opportunities for liquidation (higher price) does not come until the middle of the trading day.
Generally, the opportunities for liquidation come later than those of acquisition, which requires careful collaboration between buyers and sellers during execution and further call for multi-order optimization solutions as the fulfillment of acquisition orders depends on the cash supplied by liquidation.

\minisection{Multi-order optimization improves significantly against single-order execution}
We compare the execution details of IaC and PPO to show that jointly optimizing multiple orders is necessary to achieve high profit in multi-order execution.
Figure~(\ref{fig:overall-transaction}) shows how \IACC and PPO distribute the given order across the trading day on average.
The bars exhibit the ratio of acquisition and liquidation orders fulfilled at every minute on average, i.e., $\mathbb{E}_t[\frac{q_{t}}{M}]$.
The hollow red bars show the number of orders buyers intend to fulfill, and the solid bars show what they \textit{actually} trade considering the cash limitation.
We find from Figure~(\ref{fig:overall-transaction}) that PPO tends not to liquidate much at the beginning of the day, as it is not a good opportunity for liquidation, as shown in Figure~(\ref{fig:price-trend}), leading to slow cash replenishment.
Although PPO intends to buy many shares in the first 30 minutes of the day when the price is low, it is severely limited by a cash shortage. 
It has to postpone the acquisition operations and loses the best trading opportunities during the day.
On the contrary, \IACT coordinates both acquisition and liquidation more actively and efficiently fulfills most liquidation orders in the first hour of trading thus guaranteeing sufficient cash supply for acquisition orders, which shows the improvement on collaboration brought by optimizing all orders simultaneously through our method.

\minisection{The intended actions have been gradually refined during intention-aware communication in \IAC}
To illustrate the refine process of intended actions, we directly take $\bs{a}_{t,k}$, generated after the $k$-th round of communication ($1 \leq k \leq K=3$) from the trained policy of  \IACC and \IACT, as the final actions at each timestep $t$, and evaluate them on the test sets.
We report the profitability performance (EG) and collaboration effectiveness (TOC)
of these intended actions in Figure~(\ref{fig:diff-comm}) to exhibit the collaboration ability of the agents after each round of action refinement.
We can tell that 
(1) all intended actions achieve good TOC performance and reasonable EG, which reflects that even the intended actions proposed before the final actions have reflected the intentions of the agents thus subsequently offer clear information for better collaboration.
(2) The EG gets higher and the TOC gradually reaches optimal while the agents communicate, indicating that the agents manage to improve their intended actions based on the intentions of each other for better collaboration.
We also conduct case study on the transaction details of the intended actions after each round of communication in Appendix~\ref{app:case} to further show the refinement of intended actions during our intention-aware communication.

\begin{figure}[h]%
    \centering
    \begin{subfigure}{0.49\columnwidth}
        \centering
        \includegraphics[width=\textwidth]{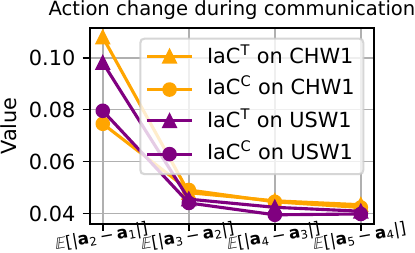}\caption{}\label{fig:delta-a}
    \end{subfigure}
    \hfill
    \begin{subfigure}{0.49\columnwidth}
        \centering
        \includegraphics[width=\textwidth]{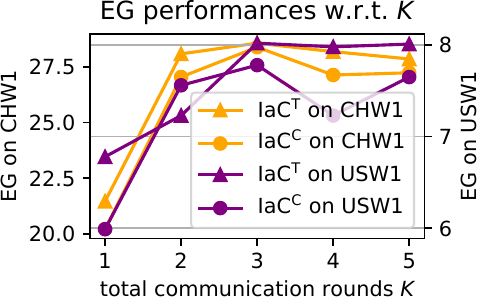}
        \caption{}\label{fig:diff_k}
    \end{subfigure}
    \vspace{-10pt}
    \caption{(a) Convergence analysis of intended actions along with communication rounds ($1 \leq k \leq K=5$); (b) Analysis of total numbers ($K$) of communication rounds.}
\end{figure}
% \vspace{-5pt}

\minisection{The intended actions have converged after multi-round communication in \IAC}
Figure~(\ref{fig:delta-a}) shows the average difference between the intended actions of neighboring rounds with totally $K=5$ communication rounds.
It indicates that the intended actions generally reaches convergence as the agents communicate for multiple rounds.
Figure~(\ref{fig:diff_k}) illustrates the influence of the total communication round number $K$ on the EG performances of our \IAC method.
The performances first improve sharply and then remain stable when $K \geq 3$, which indicates that when intention-aware communication is sufficient ($K\geq 3$), additional communication would not bring significant improvements.
These observations reflect the stability and robustness of our proposed \IAC method.

% \vspace{-4pt}
\section{Software for Order Execution}
We developed a financial decision-making toolkit \textit{Qlib.RL} based on Qlib~\cite{yang2020qlib}, to support order execution scenario.
It offers APIs for receiving orders from upstream portfolio management systems and outputs detailed execution decisions to downstream trading program.
Qlib.RL supports simultaneously execution of 1,000 orders on a machine with a NVIDIA P40 GPU and an Intel Xeon 8171M CPU, where all execution decisions would be given within 50 millisecond, which is significantly faster than the required decision time interval that is 1 minute in our practice.
As for training, Qlib.RL retrains the policy every 2 months with the latest data and applies a rolling manner to maintain promising performance, which is an acceptable cost in this scenario.
The corresponding codes and benchmark framework with data can be referred to \hyperlink{https://seqml.github.io/marl4fin}{https://seqml.github.io/marl4fin}.

\vspace{-4pt}
\section{Conclusion and Future Work}
In this paper, we formulate multi-order execution task as a multi-agent collaboration problem and solve it through an intention-aware communication method.
Specifically, we model the intention of agents as their intended actions and allow agents to share and refine their intended actions through multiple rounds of communication.
A novel action value attribution method is proposed to optimize the intended actions directly.
The experiment results have shown the superiority of our proposed method.

In the future, we plan to conduct joint optimization with order execution and portfolio management, and adapt our intention-aware communication methodology to wider RL applications.

\bibliographystyle{IEEEtran}
\balance
\bibliography{main}

\begin{thebibliography}{10}
\providecommand{\url}[1]{#1}
\csname url@samestyle\endcsname
\providecommand{\newblock}{\relax}
\providecommand{\bibinfo}[2]{#2}
\providecommand{\BIBentrySTDinterwordspacing}{\spaceskip=0pt\relax}
\providecommand{\BIBentryALTinterwordstretchfactor}{4}
\providecommand{\BIBentryALTinterwordspacing}{\spaceskip=\fontdimen2\font plus
\BIBentryALTinterwordstretchfactor\fontdimen3\font minus
  \fontdimen4\font\relax}
\providecommand{\BIBforeignlanguage}[2]{{%
\expandafter\ifx\csname l@#1\endcsname\relax
\typeout{** WARNING: IEEEtran.bst: No hyphenation pattern has been}%
\typeout{** loaded for the language `#1'. Using the pattern for}%
\typeout{** the default language instead.}%
\else
\language=\csname l@#1\endcsname
\fi
#2}}
\providecommand{\BIBdecl}{\relax}
\BIBdecl

\bibitem{wang2019alphastock}
J.~Wang, Y.~Zhang, K.~Tang, J.~Wu, and Z.~Xiong, ``Alphastock: A
  buying-winners-and-selling-losers investment strategy using interpretable
  deep reinforcement attention networks,'' in \emph{KDD}, 2019.

\bibitem{ye2020reinforcement}
Y.~Ye, H.~Pei, B.~Wang, P.-Y. Chen, Y.~Zhu, J.~Xiao, and B.~Li,
  ``Reinforcement-learning based portfolio management with augmented asset
  movement prediction states,'' in \emph{AAAI}, 2020.

\bibitem{cartea2015optimal}
A.~Cartea and S.~Jaimungal, ``Optimal execution with limit and market orders,''
  \emph{Quantitative Finance}, vol.~15, no.~8, pp. 1279--1291, 2015.

\bibitem{fang2021universal}
Y.~Fang, K.~Ren, W.~Liu, D.~Zhou, W.~Zhang, J.~Bian, Y.~Yu, and T.-Y. Liu,
  ``Universal trading for order execution with oracle policy distillation,'' in
  \emph{AAAI}, 2021.

\bibitem{almgren2001optimal}
R.~Almgren and N.~Chriss, ``Optimal execution of portfolio transactions,''
  \emph{Journal of Risk}, 2001.

\bibitem{kakade2004competitive}
S.~M. Kakade, M.~Kearns, Y.~Mansour, and L.~E. Ortiz, ``Competitive algorithms
  for vwap and limit order trading,'' in \emph{Proceedings of the 5th ACM
  conference on Electronic commerce}, 2004, pp. 189--198.

\bibitem{lee2007multiagent}
J.~W. Lee, J.~Park, O.~Jangmin, J.~Lee, and E.~Hong, ``A multiagent approach to
  $ q $-learning for daily stock trading,'' \emph{IEEE Transactions on Systems,
  Man, and Cybernetics-Part A: Systems and Humans}, vol.~37, no.~6, pp.
  864--877, 2007.

\bibitem{ning2018double}
B.~Ning, F.~H.~T. Ling, and S.~Jaimungal, ``Double deep q-learning for optimal
  execution,'' \emph{CoRR}, 2018.

\bibitem{lin2020end}
S.~Lin and P.~A. Beling, ``An end-to-end optimal trade execution framework
  based on proximal policy optimization,'' in \emph{IJCAI}, 2020.

\bibitem{zhou2019factorized}
M.~Zhou, Y.~Chen, Y.~Wen, Y.~Yang, Y.~Su, W.~Zhang, D.~Zhang, and J.~Wang,
  ``Factorized q-learning for large-scale multi-agent systems,'' in
  \emph{Proceedings of the First International Conference on Distributed
  Artificial Intelligence}, 2019, pp. 1--7.

\bibitem{tan1993multi}
M.~Tan, ``Multi-agent reinforcement learning: Independent vs. cooperative
  agents,'' in \emph{ICML}, 1993.

\bibitem{li2019cooperative}
X.~Li, J.~Zhang, J.~Bian, Y.~Tong, and T.-Y. Liu, ``A cooperative multi-agent
  reinforcement learning framework for resource balancing in complex logistics
  network,'' in \emph{Proceedings of the 18th International Conference on
  Autonomous Agents and MultiAgent Systems}.\hskip 1em plus 0.5em minus
  0.4em\relax International Foundation for Autonomous Agents and Multiagent
  Systems, 2019, pp. 980--988.

\bibitem{foerster2018counterfactual}
J.~Foerster, G.~Farquhar, T.~Afouras, N.~Nardelli, and S.~Whiteson,
  ``Counterfactual multi-agent policy gradients,'' in \emph{Proceedings of the
  AAAI Conference on Artificial Intelligence}, vol.~32, 2018.

\bibitem{zhang2018fully}
K.~Zhang, Z.~Yang, H.~Liu, T.~Zhang, and T.~Basar, ``Fully decentralized
  multi-agent reinforcement learning with networked agents,'' in
  \emph{International Conference on Machine Learning}.\hskip 1em plus 0.5em
  minus 0.4em\relax PMLR, 2018, pp. 5872--5881.

\bibitem{wang2021hierarchical}
L.~Wang, L.~Han, X.~Chen, C.~Li, J.~Huang, W.~Zhang, W.~Zhang, X.~He, and
  D.~Luo, ``Hierarchical multiagent reinforcement learning for allocating
  guaranteed display ads,'' \emph{IEEE Transactions on Neural Networks and
  Learning Systems}, 2021.

\bibitem{das2019tarmac}
A.~Das, T.~Gervet, J.~Romoff, D.~Batra, D.~Parikh, M.~Rabbat, and J.~Pineau,
  ``Tarmac: Targeted multi-agent communication,'' in \emph{International
  Conference on Machine Learning}, 2019.

\bibitem{sukhbaatar2016learning}
S.~Sukhbaatar, R.~Fergus \emph{et~al.}, ``Learning multiagent communication
  with backpropagation,'' \emph{NeurIPS}, 2016.

\bibitem{huang2020learning}
T.~Huang, Z.~Lu \emph{et~al.}, ``Learning individually inferred communication
  for multi-agent cooperation,'' \emph{NeurIPS}, 2020.

\bibitem{kim2020communication}
W.~Kim, J.~Park, and Y.~Sung, ``Communication in multi-agent reinforcement
  learning: Intention sharing,'' in \emph{ICLR}, 2020.

\bibitem{cohen1990intention}
P.~R. Cohen and H.~J. Levesque, ``Intention is choice with commitment,''
  \emph{Artificial intelligence}, 1990.

\bibitem{tomasello2005understanding}
M.~Tomasello, M.~Carpenter, J.~Call, T.~Behne, and H.~Moll, ``Understanding and
  sharing intentions: The origins of cultural cognition,'' \emph{Behavioral and
  brain sciences}, 2005.

\bibitem{hendricks2014reinforcement}
D.~Hendricks and D.~Wilcox, ``A reinforcement learning extension to the
  almgren-chriss framework for optimal trade execution,'' in \emph{CIFEr},
  2014.

\bibitem{hu2016optimal}
R.~Hu, ``Optimal order execution using stochastic control and reinforcement
  learning,'' 2016.

\bibitem{daberius2019deep}
K.~Dab{\'e}rius, E.~Granat, and P.~Karlsson, ``Deep execution-value and policy
  based reinforcement learning for trading and beating market benchmarks,''
  \emph{Available at SSRN 3374766}, 2019.

\bibitem{bertsimas1998optimal}
D.~Bertsimas and A.~W. Lo, ``Optimal control of execution costs,''
  \emph{Journal of Financial Markets}, vol.~1, no.~1, pp. 1--50, 1998.

\bibitem{nevmyvaka2006reinforcement}
Y.~Nevmyvaka, Y.~Feng, and M.~Kearns, ``Reinforcement learning for optimized
  trade execution,'' in \emph{ICML}, 2006.

\bibitem{7407387}
Y.~Deng, F.~Bao, Y.~Kong, Z.~Ren, and Q.~Dai, ``Deep direct reinforcement
  learning for financial signal representation and trading,'' \emph{IEEE
  Transactions on Neural Networks and Learning Systems}, vol.~28, no.~3, pp.
  653--664, 2017.

\bibitem{bao2019multi}
W.~Bao and X.-y. Liu, ``Multi-agent deep reinforcement learning for liquidation
  strategy analysis,'' \emph{CoRR}, 2019.

\bibitem{karpe2020multi}
M.~Karpe, J.~Fang, Z.~Ma, and C.~Wang, ``Multi-agent reinforcement learning in
  a realistic limit order book market simulation,'' \emph{Proceedings of the
  First ACM International Conference on AI in Finance}, 2020.

\bibitem{lussange2021modelling}
J.~Lussange, I.~Lazarevich, S.~Bourgeois-Gironde, S.~Palminteri, and B.~Gutkin,
  ``Modelling stock markets by multi-agent reinforcement learning,''
  \emph{Computational Economics}, 2021.

\bibitem{lussange2019stock}
J.~Lussange, I.~Lazarevich, S.~Bourgeois~Gironde, S.~Palminteri, and B.~Gutkin,
  ``Stock market microstructure inference via multi-agent reinforcement
  learning,'' \emph{CoRR}, 2019.

\bibitem{lee2020maps}
J.~Lee, R.~Kim, S.-W. Yi, and J.~Kang, ``Maps: Multi-agent reinforcement
  learning-based portfolio management system,'' in \emph{IJCAI}, 2020.

\bibitem{wang2021commission}
R.~Wang, H.~Wei, B.~An, Z.~Feng, and J.~Yao, ``Commission fee is not enough: A
  hierarchical reinforced framework for portfolio management,'' in \emph{AAAI},
  2021.

\bibitem{melo2011querypomdp}
F.~S. Melo, M.~T. Spaan, and S.~J. Witwicki, ``Querypomdp: Pomdp-based
  communication in multiagent systems,'' in \emph{European Workshop on
  Multi-Agent Systems}, 2011.

\bibitem{panait2005cooperative}
L.~Panait and S.~Luke, ``Cooperative multi-agent learning: The state of the
  art,'' \emph{Autonomous agents and multi-agent systems}, 2005.

\bibitem{foerster2016learning}
J.~Foerster, I.~A. Assael, N.~De~Freitas, and S.~Whiteson, ``Learning to
  communicate with deep multi-agent reinforcement learning,'' \emph{NeurIPS},
  2016.

\bibitem{jiang2020graph}
J.~Jiang, C.~Dun, T.~Huang, and Z.~Lu, ``Graph convolutional reinforcement
  learning,'' in \emph{ICLR}, 2020.

\bibitem{jiang2018learning}
J.~Jiang and Z.~Lu, ``Learning attentional communication for multi-agent
  cooperation,'' \emph{NeurIPS}, 2018.

\bibitem{kim2019learning}
D.~Kim, S.~Moon, D.~Hostallero, W.~J. Kang, T.~Lee, K.~Son, and Y.~Yi,
  ``Learning to schedule communication in multi-agent reinforcement learning,''
  \emph{ICLR}, 2019.

\bibitem{peng2017multiagent}
P.~Peng, Y.~Wen, Y.~Yang, Q.~Yuan, Z.~Tang, and et. al., ``Multiagent
  bidirectionally-coordinated nets: Emergence of human-level coordination in
  learning to play starcraft combat games,'' \emph{CoRR}, 2017.

\bibitem{rabinowitz2018machine}
N.~Rabinowitz, F.~Perbet, F.~Song, C.~Zhang, S.~A. Eslami, and M.~Botvinick,
  ``Machine theory of mind,'' in \emph{International conference on machine
  learning}.\hskip 1em plus 0.5em minus 0.4em\relax PMLR, 2018.

\bibitem{raileanu2018modeling}
R.~Raileanu, E.~Denton, A.~Szlam, and R.~Fergus, ``Modeling others using
  oneself in multi-agent reinforcement learning,'' in \emph{ICML}, 2018.

\bibitem{lowe2017multi}
R.~Lowe, Y.~I. Wu, A.~Tamar, J.~Harb, O.~Pieter~Abbeel, and I.~Mordatch,
  ``Multi-agent actor-critic for mixed cooperative-competitive environments,''
  \emph{NeurIPS}, 2017.

\bibitem{cartea2015algorithmic}
{\'A}.~Cartea, S.~Jaimungal, and J.~Penalva, \emph{Algorithmic and
  high-frequency trading}.\hskip 1em plus 0.5em minus 0.4em\relax Cambridge
  University Press, 2015.

\bibitem{schulman2017proximal}
J.~Schulman, F.~Wolski, P.~Dhariwal, A.~Radford, and O.~Klimov, ``Proximal
  policy optimization algorithms,'' \emph{CoRR}, 2017.

\bibitem{jegadeesh1993returns}
N.~Jegadeesh and S.~Titman, ``Returns to buying winners and selling losers:
  Implications for stock market efficiency,'' \emph{The Journal of finance},
  1993.

\bibitem{yang2020qlib}
X.~Yang, W.~Liu, D.~Zhou, J.~Bian, and T.-Y. Liu, ``Qlib: An ai-oriented
  quantitative investment platform,'' \emph{CoRR}, 2020.

\bibitem{bhattacharya2002median}
B.~Bhattacharya and D.~Habtzghi, ``Median of the p value under the alternative
  hypothesis,'' \emph{The American Statistician}, 2002.

\bibitem{cho2014learning}
K.~Cho, B.~van Merrienboer, C.~Gulcehre, D.~Bahdanau, F.~Bougares, H.~Schwenk,
  and Y.~Bengio, ``Learning phrase representations using rnn encoder--decoder
  for statistical machine translation,'' in \emph{EMNLP}, 2014.

\bibitem{vaswani2017attention}
A.~Vaswani, N.~Shazeer, N.~Parmar, J.~Uszkoreit, L.~Jones, A.~N. Gomez,
  L.~Kaiser, and I.~Polosukhin, ``Attention is all you need,'' \emph{NeurIPS},
  2017.

\end{thebibliography}

\newpage
\appendix

\section{Detailed Method Settings}
We provide descriptions of some essential settings of the MDP design and method implementation in this section.

\subsection{Statistics of datasets}\label{app:dataset}
All the compared methods are trained and evaluated based on the historical data of China A-share stock market and US stock market.
And we conduct a rolling window based data division for robust and fair evaluation.
The detailed statistic of all datasets are presented in Table~\ref{tab:dataset-statistics}

\begin{table}[h]
    \centering
	\resizebox{\linewidth}{!}{\large
		\begin{tabular}{|c|c|l|r|c|}\hline
        Market & Rolling window             & \multicolumn{1}{c|}{Phase} & \multicolumn{1}{c|}{\# orders} & Time period            \\ \hline
        \multirow{9}{*}{\begin{tabular}[c]{@{}c@{}}China A-share \\ Stock Market\end{tabular}} & \multirow{3}{*}{CHW1} & Training                   & 910,700                       & 01/01/2018 - 31/08/2018  \\ \cline{3-5} 
                                  & & Validation                 & 19,740                        & 01/09/2018 - 31/10/2018  \\ \cline{3-5} 
                                  & & Test                       & 22,480                        & 01/11/2018 - 31/12/2018 \\ \cline{2-5}
        & \multirow{3}{*}{CHW2} & Training                   & 966,020                       & 01/01/2019 - 31/08/2019   \\ \cline{3-5} 
                                  & & Validation                 & 22,820                        & 01/09/2019 - 31/10/2019  \\ \cline{3-5} 
                                  & & Test                       & 25,220                        & 01/11/2019 - 31/12/2019 \\ \cline{2-5}
        &\multirow{3}{*}{CHW3} & Training                   & 1,127,220                       & 01/03/2020 - 31/10/2020  \\ \cline{3-5} 
        &                           & Validation                 & 26,420                        & 01/11/2020 - 31/12/2020  \\ \cline{3-5} 
        &                           & Test                       & 20,260                       & 01/01/2021 - 28/02/2021 \\
                                \hline\hline
        \multirow{9}{*}{US Stock Market} & \multirow{3}{*}{USW1} & Training                   & 800,020                       & 01/01/2018 - 31/08/2018   \\ \cline{3-5} 
        &                           & Validation                 & 18,330                    	& 01/09/2018 - 31/10/2018  \\ \cline{3-5} 
        &                           & Test                       & 17,240                    	& 01/11/2018 - 31/12/2018 \\ \cline{2-5}
        & \multirow{3}{*}{USW2} 		& Training                   & 827,920                   	& 01/01/2019 - 31/08/2019   \\ \cline{3-5} 
        &                           & Validation                 & 17,540                      	& 01/09/2019 - 31/10/2019  \\ \cline{3-5} 
        &                           & Test                       & 18,820                      	& 01/11/2019 - 31/12/2019 \\ \cline{2-5}
        & \multirow{3}{*}{USW3} 		& Training                   & 900,020                    & 01/03/2020 - 31/10/2020   \\ \cline{3-5} 
        &                           & Validation                 & 17,230                      	& 01/11/2020 - 31/12/2020  \\ \cline{3-5} 
        &                           & Test                       & 18,440                      	& 01/01/2021 - 28/02/2021 \\ \hline
        \end{tabular}
	}
    \caption{The dataset statistics on two real-world markets.
	}\label{tab:dataset-statistics}
\vspace{-30pt}
\end{table}
\subsection{Details of state space}\label{app:state}
The state $\s_t \in \mathcal{S}$ describes the overall information of the whole system, and each agent can observe the \textit{private} information related to the target asset of the corresponding order
and some \textit{public} information shared by all agents.
The detailed contents of both parts are listed in Table~\ref{tab:state}.

\begin{table}[h]
\resizebox{.88\columnwidth}{!}{
\large
\begin{tabular}{|c|l|}
\hline
Category                 & Description                                                      \\ \hline
\multirow{2}{*}{Public} & elapsed timestep normalized by time horizon length as $t$ / $T$                                            \\ \cline{2-2} 
                         & history of cash balance $(c_0, c_1, \ldots, c_t)$     \\ \hline
\multirow{4}{*}{Private} & history of the traded volume of order $i$ $(q_1^i, \ldots, q_t^i)$      \\ \cline{2-2} 
                         & history of market information, i.e., asset price and \\ 
                        %  \cline{2-2}
                         & transaction volume, of the corresponding asset of order $i$\\ \cline{2-2} 
                         & the trading direction $d^i \in \{1, -1\}$                           \\ \hline
\end{tabular}}
\caption{Information in state $\s_t^i$ for agent $i$ at timestep $t$.}
\label{tab:state}
\end{table}

\begin{table*}[h]
\resizebox{.9\textwidth}{!}{
\begin{tabular}{|c|crrrrr|crrrrr|}
\hline
Dataset & \multicolumn{6}{c|}{China A-share Market Window 1 (CHW1)} & \multicolumn{6}{c|}{US Stock Market Window 1 (USW1)} \\ \hline
Action space & \multicolumn{2}{c|}{\{0, 0.33, 0.67, 1\}} & \multicolumn{2}{c|}{\{0, 0.25, 0.5, 0.75, 1\}} & \multicolumn{2}{c|}{\{0, 0.2, 0.4, 0.6, 0.8, 1\}} & \multicolumn{2}{c|}{\{0, 0.33, 0.67, 1\}} & \multicolumn{2}{c|}{\{0, 0.25, 0.5, 0.75, 1\}} & \multicolumn{2}{c|}{\{0, 0.2, 0.4, 0.6, 0.8, 1\}} \\ \hline
Metric & EG (\textpertenthousand) $\uparrow$ & \multicolumn{1}{c|}{TOC (\%) $\downarrow$} & \multicolumn{1}{c}{EG (\textpertenthousand) $\uparrow$} & \multicolumn{1}{c|}{TOC (\%) $\downarrow$} & \multicolumn{1}{c}{EG (\textpertenthousand) $\uparrow$} & \multicolumn{1}{c|}{TOC (\%) $\downarrow$} & EG (\textpertenthousand) $\uparrow$ & \multicolumn{1}{c|}{TOC (\%) $\downarrow$} & \multicolumn{1}{c}{EG (\textpertenthousand) $\uparrow$} & \multicolumn{1}{c|}{TOC (\%) $\downarrow$} & \multicolumn{1}{c}{EG (\textpertenthousand) $\uparrow$} & \multicolumn{1}{c|}{TOC (\%) $\downarrow$} \\ \hline
PPO & \multicolumn{1}{r}{23.33$\pm$5.23} & \multicolumn{1}{r|}{48.23$\pm$0.22} & 21.63$\pm$1.45 & \multicolumn{1}{r|}{40.39$\pm$6.66} & 20.87$\pm$0.98 & 37.23$\pm$5.67 & \multicolumn{1}{r}{4.66$\pm$1.07} & \multicolumn{1}{r|}{15.57$\pm$3.42} & 4.02$\pm$0.46 & \multicolumn{1}{r|}{14.18$\pm$3.43} & 3.99$\pm$0.32 & 12.19$\pm$3.23 \\ \hline
TarMAC & \multicolumn{1}{r}{23.19$\pm$1.98} & \multicolumn{1}{r|}{4.34$\pm$0.52} & 22.46$\pm$1.42 & \multicolumn{1}{r|}{2.56$\pm$0.10} & 21.18$\pm$1.45 & 1.99$\pm$0.23 & \multicolumn{1}{r}{6.95$\pm$1.35} & \multicolumn{1}{r|}{2.97$\pm$0.78} & 6.54$\pm$1.05 & \multicolumn{1}{r|}{2.34$\pm$0.68} & 6.57$\pm$1.15 & 2.26$\pm$1.23 \\ \hline
\IACT & \multicolumn{1}{r}{28.99$\pm$3.58} & \multicolumn{1}{r|}{2.34$\pm$0.54} & 28.36$\pm$3.45 & \multicolumn{1}{r|}{1.23$\pm$0.36} & 27.12$\pm$2.34 & 1.22$\pm$0.23 & \multicolumn{1}{r}{8.25$\pm$0.56} & \multicolumn{1}{r|}{1.38$\pm$0.41} & 8.11$\pm$0.52 & \multicolumn{1}{r|}{1.01$\pm$0.22} & 7.96$\pm$0.47 & 0.99$\pm$0.41 \\ \hline
\end{tabular}
}
\caption{The experiment results of different action space settings. $\uparrow$ ($\downarrow$) means higher (lower) value is better.
}\label{tab:action-space}
\end{table*}

\subsection{Multi-order execution procedure}\label{app:trading}
\begin{figure}[t]
	\centering
		\includegraphics[width=0.8\columnwidth]{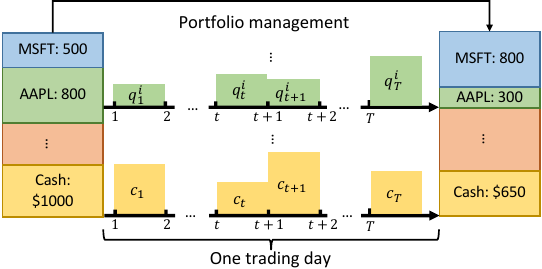}
	\caption{The detailed multi-order execution procedure during one trading day.
	}\label{fig:eva_protocal}
\end{figure}

We describe the detailed multi-order execution procedure of each day in this section.
The procedure is illustrated in Figure~\ref{fig:eva_protocal}.
The orders to execute, consisting of the assets, the amount to trade and the trading directions of the orders, are first given by an upstream portfolio management strategy.
All orders should be fulfilled before the time horizon ends, i.e., within one trading day.
Note that, we assume it possible for all orders to be fully fulfilled before the end of the time horizon, which is usually guaranteed by the portfolio management strategy. 
The trading day is divided evenly into $T$ timesteps in total.
For the $i$-th asset, at the \textit{beginning} of timestep $t$, the state $\s_t^i$ is generated by the environment and fed into the agent policy $\pi^i$, and the corresponding action $a^i_t$ is proposed by the policy.
The proposed volume to execute $q^i_{t} = a^i_t M^i$ calculated based on action $a^i_t$ will be executed at this timestep $t$.
Also, the cash consumed by all acquisition operations and replenished by liquidation operations is calculated at each timestep to update the cash balance $c_t$.

Specifically, in all of our experiments, we set the time for each period as 30 minutes, thus $T=8$ for China A-share stock market and $T=13$ for US stock market.
Without loss of generality, following \cite{ning2018double,fang2021universal}, within each timestep, we use TWAP as a lower level strategy to conduct actual execution at each minute, in other words, we equally allocate some volume on every minute in timestep $t$ from $q^i_{t}$.
Note that, one can also replace TWAP with any other order execution strategies.

\subsection{Estimation of Additional Annualized Rate of Return}\label{app:arr}
We calculate additional annualized rate of return (ARR) brought by order-execution strategies relative to TWAP execution strategy under the same portfolio management strategy as evaluation.
ARR is formally estimated as 
\begin{equation}
\begin{aligned}
\text{ARR} \approx & \left[ \left( 1 + 
\text{EG} \times \text{daily turnover rate} \right) ^{(\text{num. of trading days per year})} \right. \\
&~~~~~~~~~~~~~~~~ \left. - 1 \right] \times 100\%~,
\end{aligned}
\end{equation}
where we take daily turnover rate as 10\% and 250 trading days per year for both Chinese and US stock market.

\subsection{Hyper-parameter Settings}\label{app:para}
In our experiment, we set the discount factor $\gamma = 1$, the coefficient for the penalties $R_a^{-}$ and $R_c^{-}$ in reward function defined in Eq.~(\ref{eq:rew2}) and Eq.~(\ref{eq:rew3}) are set to $\alpha=0.01$ and $\sigma=\frac{1}{30}$, respectively,
The hyper-parameters of all RL methods are listed in Table \ref{tab:param}, together with the searching ranges within which we searched the best hyper-parameters on the validation set.

\begin{table}[h]
\resizebox{\columnwidth}{!}{
\begin{tabular}{|c|ccccccc|c|}
\hline
Parameter & \multicolumn{1}{c|}{PPO} & \multicolumn{1}{c|}{DDQN} & \multicolumn{1}{c|}{CommNet} & \multicolumn{1}{c|}{TarMAC} & \multicolumn{1}{c|}{IS} & \multicolumn{1}{c|}{\IACC} & \IACT & Search space \\ \hline
Learning rate & \multicolumn{7}{c|}{1e-4} & \{1e-4,5e-5,1e-5\} \\ \hline
Batch size & \multicolumn{7}{c|}{128} & \{64, 128\} \\ \hline
RNN cell size & \multicolumn{7}{c|}{64} & \{64,128,256\} \\ \hline
FC-layer size & \multicolumn{7}{c|}{128} & \{64, 128, 256\} \\ \hline
Optimizer & \multicolumn{7}{c|}{Adam} & ---- \\ \hline
Target network update frequency & \multicolumn{1}{c|}{-} & \multicolumn{1}{c|}{200} & \multicolumn{1}{c|}{-} & \multicolumn{1}{c|}{-} & \multicolumn{1}{c|}{-} & \multicolumn{1}{c|}{-} & - & \{100,200\} \\ \hline
Policy network update frequency & \multicolumn{1}{c|}{2000} & \multicolumn{1}{c|}{16} & \multicolumn{1}{c|}{2000} & \multicolumn{1}{c|}{2000} & \multicolumn{1}{c|}{2000} & \multicolumn{1}{c|}{2000} & 2000 & \{16, 256, 1000, 2000\} \\ \hline
GAE $\lambda$ & \multicolumn{1}{c|}{0.9} & \multicolumn{1}{c|}{-} & \multicolumn{1}{c|}{0.9} & \multicolumn{1}{c|}{0.9} & \multicolumn{1}{c|}{0.9} & \multicolumn{1}{c|}{0.9} & 0.9 & \{0.9, 0.95\} \\ \hline
\# Attention heads & \multicolumn{1}{c|}{-} & \multicolumn{1}{c|}{-} & \multicolumn{1}{c|}{-} & \multicolumn{1}{c|}{4} & \multicolumn{1}{c|}{4} & \multicolumn{1}{c|}{-} & 2 & \{2, 4\} \\ \hline
\# Communication rounds & \multicolumn{1}{c|}{-} & \multicolumn{1}{c|}{-} & \multicolumn{1}{c|}{2} & \multicolumn{1}{c|}{3} & \multicolumn{1}{c|}{3} & \multicolumn{1}{c|}{3} & 3 & \{1,2,3,4,5\} \\ \hline
\end{tabular}
}
\caption{Hyper-parameters and the search range of hyper-parameter tuning for each algorithm.
}\label{tab:param}
\end{table}

\begin{figure}
    \centering
    \includegraphics[width=\columnwidth]{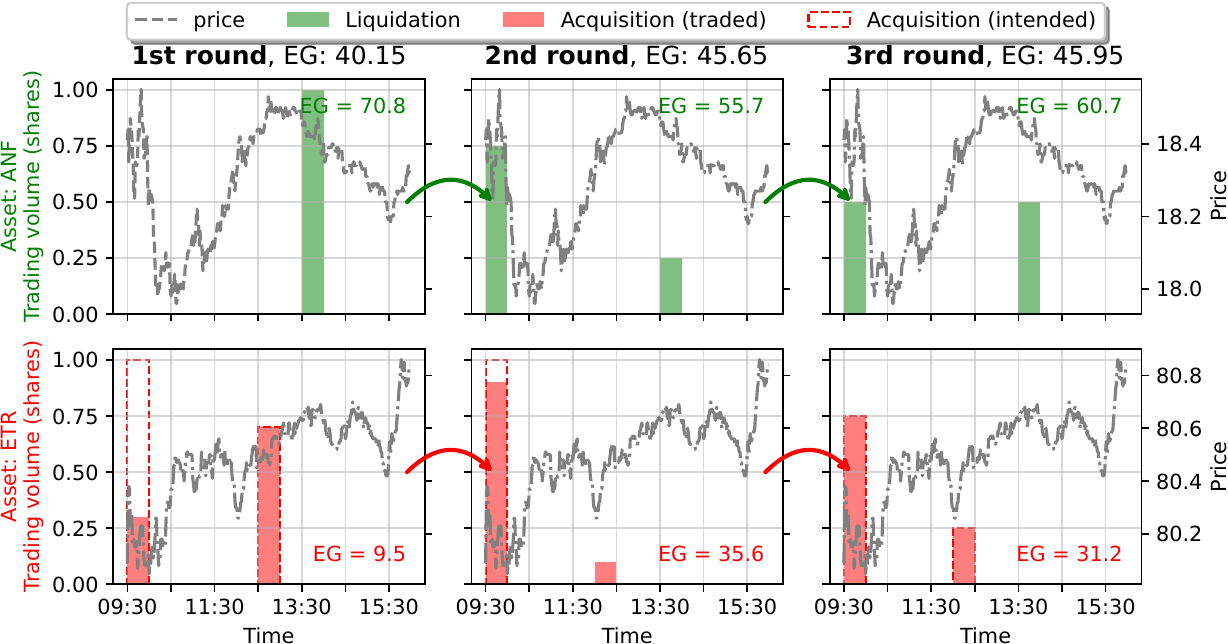}
    \caption{Case study: the execution details of intended actions after different round of communication (\IACT) between ANF (liquidation) and ETR (acquisition) on 2018-01-08 in US stock market.
    }
    \label{fig:case-study-us}
\end{figure}
\subsection{Network architecture}\label{app:network}
\begin{figure}[t]
  \begin{center}
   \includegraphics[width=0.55\columnwidth]{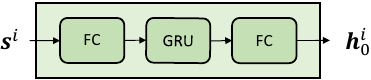}
  \end{center}
  \caption{The overall structure of the extractor network $E$. The hidden state before the first round of communication $\bs{h}_0^i$ is generated by a temporal extractor from the observation $\s^i$.
    }\label{fig:extractor} 
\end{figure}

Some specific details of the network structure of the compared methods are explained here, including the extractor, the communication channel, the decision module and the value function estimator.

\minisection{Extractor $E(\cdot)$} 
As illustrated in Figure~\ref{fig:extractor}, the extractor network $E$ is designed to extract the information in the observation of agent $i$ into an embedding vector $\bs{h}_0^i$.
As the observation $\s^i$ consists of sequences of historical market information, we encode it with a temporal extractor, which is composed by a Gated Recurrent Unit \cite{cho2014learning} (GRU) and two fully-connect (FC) layers with ReLU $\lambda(x) = max(0, x)$ as the activation function.
The structure of extractor is shared by all RL-based compared methods.

\minisection{Communication channel $C(\cdot)$}
As we mentioned before, our intention-aware communication method does not require specific form of communication channel and can be combined with arbitrary communication implementations in previous works.
In this paper, we follow \cite{foerster2016learning,das2019tarmac} and use fully connected network and self-attention module \cite{vaswani2017attention} as communication channel.
Although graph neural networks (GNN) have been used as communication channels in recent works~\cite{huang2020learning,jiang2020graph,jiang2018learning}, which is also suitable in our method,
we did not utilize GNN as there is no clear graph structure between agents in multi-order execution.
Nevertheless, GNN can also be incorporated with our methods in other scenarios.

\minisection{Decision module  $D(\cdot)$}
Receiving $\bs{h}^i$ after communication, a three-layer MLP with ReLU activation and Softmax $\sigma(\bs{x})_j=\frac{e^{x_j}}{\Sigma_{m=1}^{M}e^{x_{m}}}$ as the last activation function is used to generate action distribution $\bs{\pi}^i = D(\bs{h}^i)$.

\minisection{Value function}
We use the hidden representation $\bs{h}_0$ generated by the extractor $E(\cdot)$ together with the actions $\bs{a}$ to estimate the action value.
$\bs{h}_0$ and $\bs{a}$ are fed into a three-layer MLP to derive the action value estimation $\hat{Q}(\s_t, \bs{a})$.

To improve scalability and flexibility of order execution with various order number,
all agents share the same network parameter with each other
which is a common approach in 
many previous MARL works~\cite{jiang2018learning,foerster2016learning}.
Each experiment has been conducted on the machine with an NVIDIA P40 GPU and an Intel Xeon 8171M CPU within 170 hours.
And our proposed \IAC methods, with a much faster convergence speed, achieve the best performance within 35 hours of training, as shown in Appendix~\ref{sec:lc}.

\section{In-depth Analysis}
 
\subsection{Learning Analysis}~\label{sec:lc}
\begin{figure}
    \centering
    \begin{subfigure}{0.495\columnwidth}
        \centering
        \includegraphics[width=\textwidth]{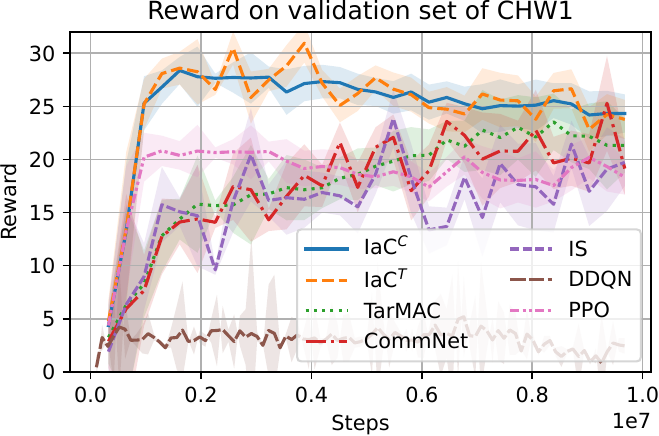}
    \end{subfigure}
    \hfill
    \begin{subfigure}{0.495\columnwidth}
        \centering
        \includegraphics[width=\textwidth]{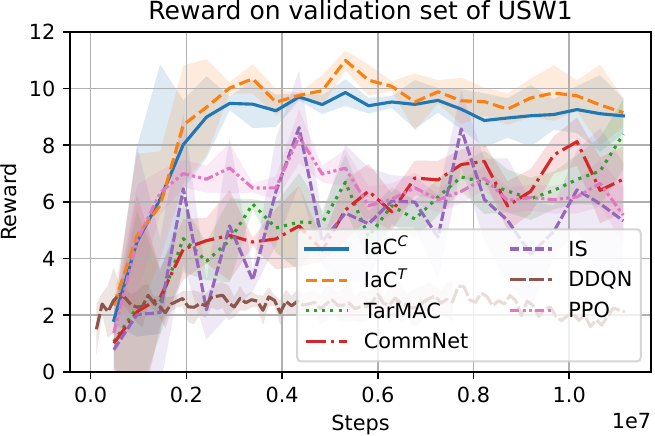}
    \end{subfigure}
    \caption{The learning curves (mean $\pm$ standard deviation of performance over six random seeds) on CHW1 and USW1.
    Here ``step'' means policy interaction with the environment.
    }\label{fig:lc}
\end{figure}

We illustrated the learning situation of the compared methods on valid sets of CHW1 and USW1 in Figure~\ref{fig:lc}.
We can tell that, (1) with our intention-aware communication method, both \IACC and \IACT converge fastest achieving the highest reward among all the compared methods.
(2) The convergence speeds of PPO and DDQN are quick, as they are optimized for single-order execution, which is less complicated thus easier to optimize than multi-order execution optimization.
However, the final performances are limited and suboptimal due to lacking of collaboration.
(3) The performance of IS has larger variance on multiple runs than TarMAC and CommNet, which might be the result of the unstable prediction of the imagined trajectories under noisy financial scenario.

\subsection{Influence of MDP settings}\label{app:analysis}

We investigate the influence of MDP settings to the trading performance.
We take action space $\mathcal{A}$ as an example and test different action space settings for PPO, TarMAC and \IACT on CHW1 and USW1.
Specifically, apart from the action space defined in Sec. \ref{sec:mdp}, we conduct experiments on other two action space settings.
The results are presented in Table~\ref{tab:action-space}.

We can tell from the results that (1) on all these action space settings, \IACT achieves the highest EG and lowest TOC, which shows that the performance improvement brought by our \IAC method is consistent.
(2) When the size of action space gets smaller, e.g., from \{0, 0.25, 0.5, 0.75, 1\} to \{0, 0.33, 0.67, 1\}, the strategies tend to achieve higher average EG performances.
However, the performances get more unstable, and the TOC results get worse (higher).
It is understandable as the agents have to trade more at one timestep, thus having a larger chance to trade much at a good opportunity of the day while suffering from large variance.
In the meantime, the agents fail to conduct fine-grained execution and collaboration, resulting in worse, i.e., higher, TOC results.
\begin{table}[h]
\centering
\resizebox{0.7\columnwidth}{!}{
\begin{tabular}{|c|c|c|c|c|c|c|}
\hline
 & CHW1 & CHW2 & CHW3 & USW1 & USW2 & USW3 \\ \hline
AV (\textperthousand) & 7.18 & 7.18 & 7.93 & 4.22 & 4.92 & 4.66\\ \hline
ASM (\textpertenthousand) & 0.71 & 0.89 & 0.87 & 0.29 & 0.34 & 0.27 \\ \hline
\end{tabular}}
\caption{The average volatility and momentum of all datasets.}\label{tab:market}
\end{table}
\subsection{Influences of market situation}\label{app:market}
There exists a significant gap between the EG results on China A-share stock market and US stock market for all the compared methods.
We conduct analysis on the overall market situation of these two markets to explain this phenomenon.
Specifically, we calculate the average volatility (AV) and average strength of momentum (ASM) of stock prices on each trading day as
$\text{AV} = \frac{1}{|\mathbbm{D}|}\sum_{i=1}^{|\mathbbm{D}|}\text{std}(\frac{p^i_t}{\tilde{p}^i})~,$
$\text{ASM} = \frac{1}{|\mathbbm{D}|}\sum_{i=1}^{|\mathbbm{D}|}\sum_{t=1}^{T-1}|p^i_t - p^i_{t-1}|~,$
where std means standard deviation, $|\mathbbm{D}|$ is the number of orders in the dataset and $\tilde{p}^i$ is the average market price of asset $i$ we defined in Eq.~(\ref{eq:aep}).

The statistics are shown in Table~\ref{tab:market}.
We can see that both the ASM and the AV of datasets of China A-share market are larger than that of US stock market, indicating that the price movements on Chinese stocks tends to be larger and have more obvious trends, which makes it easier for RL policies to find good trading opportunities.

\subsection{Case study}\label{app:case}
Figure~\ref{fig:case-study-us} illustrates the detailed trading situation of intended action after different rounds of communication of \IACT on USW1 in one trading day.
It shows that the agents sacrifice the profit of liquidation orders a little, to trade earlier for cash supply to maximize the overall profit of both acquisition and liquidation.
The agents manages to generally refine their actions and achieve better collaboration through multiple rounds of intention-aware communication.
From Figure~\ref{fig:case-study-us}, the case study clearly presents the collaborative effectiveness of the learned policy, which has illustrated the efficacy of our proposed method.

\end{document}